\pdfoutput=1

\documentclass[11pt]{article}

\usepackage[]{EMNLP2023}

\usepackage{times}
\usepackage{latexsym}
\usepackage{graphicx}
\usepackage{xcolor}
\usepackage{colortbl}
\usepackage{amsmath}
\usepackage{tcolorbox}
\usepackage{booktabs}
\usepackage{array}
\usepackage{multirow}
\usepackage{subcaption}
\usepackage{setspace}

\definecolor{bggreen}{rgb}{0.902 0.984 0.902}
\definecolor{bgblue}{rgb}{0.890 0.996 1.000}
\definecolor{bgpink}{rgb}{1.000 0.886 0.976}
\definecolor{bgyellow}{rgb}{1.000 0.988 0.851}

\usepackage[T1]{fontenc}

\usepackage[utf8]{inputenc}

\usepackage{microtype}

\usepackage{inconsolata}

\title{Exploring Large Language Models for Communication Games: \\An Empirical Study on Werewolf}

\author{\textbf{Yuzhuang Xu}$^{1}$, \textbf{Shuo Wang}$^{1}$, \textbf{Peng Li}$^{2,*}$, \textbf{Fuwen Luo}$^{1}$\\ \textbf{Xiaolong Wang}$^{1}$, \textbf{Weidong Liu}$^{1,3}$, \textbf{Yang Liu}$^{1,2,}$\thanks{$^{*}$Corresponding authors} \\
$^{1}$Department of Computer Science \& Technology, Tsinghua University, Beijing, China \\
$^{2}$Institute for AI Industry Research (AIR), Tsinghua University, Beijing, China \\
$^{3}$Zhongguancun Laboratory, Beijing, China \\
\texttt{xyz21thu@gmail.com, lipeng@air.tsinghua.edu.cn} \\
\texttt{liuyang2011@tsinghua.edu.cn}
}

\begin{document}
\maketitle
\begin{abstract}
  Communication games, which we refer to as incomplete information games that heavily depend on natural language communication, hold significant research value in fields such as economics, social science, and artificial intelligence. In this work, we explore the problem of how to engage large language models (LLMs) in communication games, and in response, propose a tuning-free framework. Our approach keeps LLMs frozen, and relies on the retrieval and reflection on past communications and experiences for improvement. An empirical study on the representative and widely-studied communication game, ``Werewolf'', demonstrates that our framework can effectively play Werewolf game without tuning the parameters of the LLMs. 
  More importantly, strategic behaviors begin to emerge in our experiments, suggesting that it will be a fruitful journey to engage LLMs in communication games and associated domains.


\end{abstract}

\section{Introduction}
\label{sec:intro}

Since incomplete information games such as Werewolf~\citep{mafia2022} and Poker~\citep{brown2019poker}  can be used as a good proxy to exploit various fundamental problems in economics and social science~\cite{gibbons1992primer}, research on playing such games with artificial intelligence (AI) agents has attracted widespread attention in recent years~\citep{brown2019poker,FAIR2022diplomacy,Toriumi2017AIWolf}.
Among them, the communication games which heavily rely on natural language communication, e.g., Werewolf, present even greater practical values and challenges as agents must gather and infer information from the inherently ambiguous natural language utterances.
Although substantial efforts have been devoted to such games~\citep{Toriumi2017AIWolf,FAIR2022diplomacy}, most of them either impose strict restrictions on the language used in the game~\citep{rule2014, werewolf2016, rl2023} or require a significant amount of human-annotated data~\citep{FAIR2022diplomacy,kramar2022negotiation}.
Therefore, it is still challenging for AI agents to play communication games in a natural way.

Fortunately, large language models (LLMs) like ChatGPT~\citep{chatgpt2022} have recently made significant advancements. 
These models have demonstrated impressive or even superhuman performance across a broad spectrum of academic and professional exams~\cite{openai2023gpt4}, showcasing sophisticated language comprehension, generation, and reasoning abilities. 
Furthermore, studies have shown that LLMs exhibit a certain degree of theory of mind capabilities~\citep{bubeck2023sparks,shapira2023clever,kosinski2023theory}, as well as the potential to simulate believable human behaviors~\cite{town2023}.
Recent research also suggests that LLMs can improve themselves~\citep{negotiation2023} or align better with human values~\citep{liu2023training} through mutual communication.
All these advancements make LLMs promising candidates for tackling the challenge of enabling AI agents to participate in communication games in a more natural and sophisticated manner.

Nevertheless, it is not trivial to play communication games for LLMs.
Firstly, the finite maximum input length of LLMs, also known as context length, limits the volume of information that can be conveyed at a single time. In communication games, historical information is important for decision-making, but it is often too massive to be processed by LLMs.
{\color{red}}Secondly, understanding the intentions of other players and making suitable decisions to win the game require complex reasoning, which is a demanding task for LLMs~\citep{least2022}.
Thirdly, LLMs might learn from experience like human beings to upgrade their behaviors. Unfortunately, fine-tuning LLMs is not practical since it is both time-consuming and data-intensive.

\begin{figure*}[ht]
  \centering
  \includegraphics[width=0.99\textwidth]{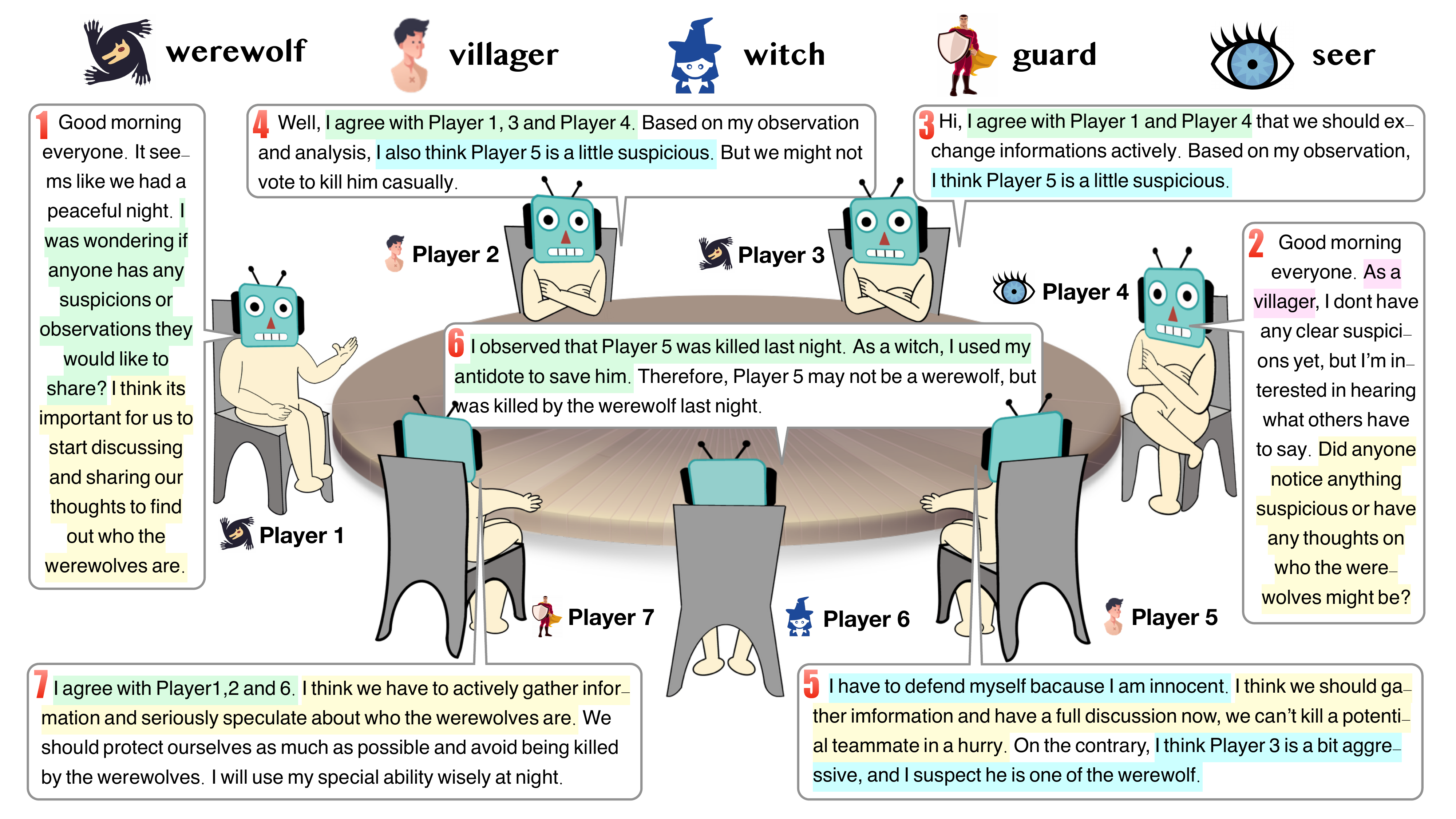}
  \caption{A snapshot of our implemented Werewolf game. There are 5 roles and 7 players, and each of them is acted by an LLM autonomously. The number before each talking denotes the speaking order. Some social behaviors can be primarily observed in this figure, including \colorbox{bggreen}{\begin{minipage}[c][0.65\baselineskip]{1.7em}trust\end{minipage}}, \colorbox{bgblue}{\begin{minipage}[c][0.65\baselineskip]{5.2em}confrontation\end{minipage}}, \colorbox{bgpink}{\begin{minipage}[c][0.65\baselineskip]{4.5em}camouflage\end{minipage}}, and \colorbox{bgyellow}{\begin{minipage}[c][0.65\baselineskip]{4.0em}leadership\end{minipage}}.}
\label{fig:werewolf}
    \vspace{-1em}
\end{figure*}

In this work, we aim to explore LLM-based agents for the Werewolf game, which is a representative and widely studied communication game. To address the issue of limited context length, we propose a method to retrieve and reflect necessary historical information, resulting in a compact context for each LLM-based agent.
{\color{red}}Moreover, the reflection process also serves the purpose of enhancing the reasoning ability of the agent, which functions in a manner akin to the chain-of-thought mechanism~\citep{cot2022}.
To learn from experience without tuning model parameters on supervised data, we propose a mechanism that extracts suggestion from past experiences based on the current situation. Our goal is to prevent LLMs from making similar mistakes repeatedly across several matches. Experiments indicate that LLMs have great potential in playing communication games. Our contributions can be summarized as follows:

\begin{itemize}
  \item We propose a framework for playing communication games with frozen LLMs without human-annotated data.
  \item {\color{red}}Empirical studies on Werewolf demonstrate that our framework demonstrates the ability to learn from experiences without tuning the parameters of LLMs.
  \item Strategic behaviors such as trust, confrontation, camouflage, and leadership begin to emerge in our experiments, which can serve as a catalyst for further research on LLMs for communication games.
\end{itemize}

\section{Background: Werewolf}

There are various versions of the Werewolf game. 
Fig.~\ref{fig:werewolf} shows an example of the version that we adopt in this work.
Specifically, there are seven players with five distinct roles: two werewolves, two villagers, a witch, a guard, and a seer.
All the involved roles are divided into two sides, of which one side is the werewolves and the other side includes the villagers and the special roles (i.e., witch, guard, and seer).
The objective of werewolves is to eliminate all villagers, while the villagers aim to work with special roles to eliminate all werewolves. There should be at least one alive villager at the end of the game if the villagers and special roles want to win.
The game alternates between day and night phases. During each night, the werewolves can vote to eliminate one role. During the daytime, all alive players will organize an open discussion and then vote to eliminate one suspicious werewolf.
As for the special roles, the witch can use a bottle of antidote and a bottle of poison, which can be used only once in a game, to either save or poison a role. The guard can protect one role to be not eliminated each night. And the seer can uncover the role of one player each night.

One important feature of the Werewolf game is that all the players only know their own roles at the beginning. They have to infer the roles of other players through natural language-based communication and reasoning.
Therefore, to excel at Werewolf, an agent should not only be good at natural language understanding and generation but also possess advanced abilities, such as deciphering the intentions of others and understanding the theory of mind~\cite{Toriumi2017AIWolf}. This factor makes Werewolf a good testbed for research on communication games.

\section{Playing Werewolf with LLMs}

\subsection{Notations}
We refer to one full day-night cycle as one {\bf day}, indexed by $t$.
A {\bf round} consists of multiple days, from the beginning of the game to the day that one side wins or it reaches the predefined max number of days. We will index a round by $r$.
The agents are numbered by $i$. In the following sections, a symbol in the form $X_i^{(r,t)}$ means it is corresponding to agent $i$ at round $r$ and day $t$. For brevity, $r$ or $t$ will be omitted when it is clear from the context.
The words an agent says to others are called {\bf responses} and the words an agent hears are called {\bf observations}, denoted as $G$ and $O$. Moreover, the agent will also generate natural language summary of the current situation given the communication history, which is called {\bf reflection} and denoted as $R$ (see \S\ref{sec:convmodeling} for more information). For brevity, we will refer to responses, observations, and reflections as {\bf messages} if they need to be considered together.

\subsection{Overall Framework}

\begin{figure}
    \centering
    \includegraphics[width=0.48\textwidth]{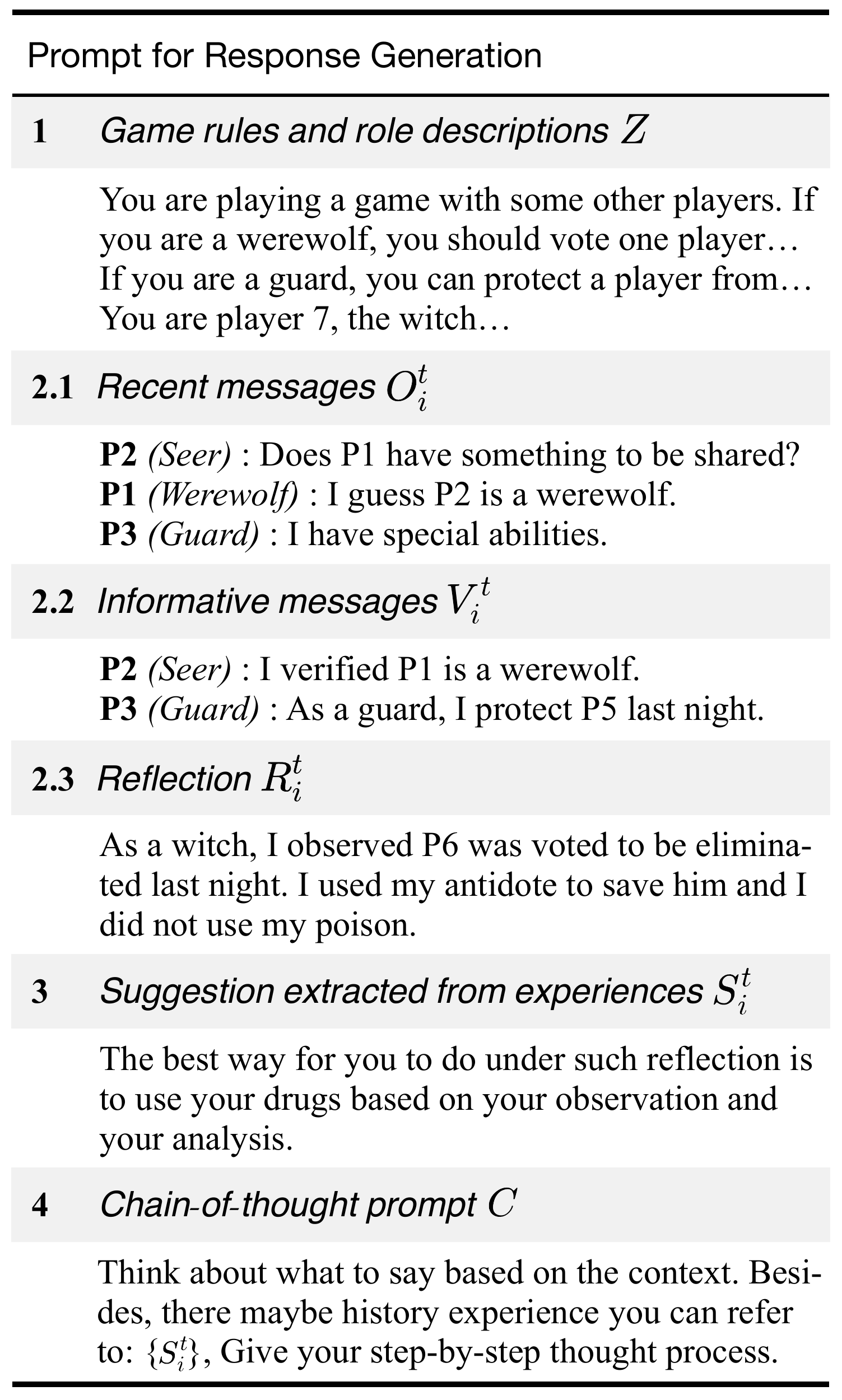}
    \caption{Outline of prompt for response generation. {\it Italics} are comments.}
    \label{fig:prompt}
    \vspace{-1em}
\end{figure}

For each role in the game, we implement an individual LLM-based agent through prompting and the full prompt can be found in {\color{red}}Appendix~\ref{app:prompt}.
Fig.~\ref{fig:prompt} shows the outline of the prompt for response generation, which consists of four major components: (1) the game rules, the assigned role, the abilities and objectives of each role, and some basic human priors on effective gameplay strategies (part 1); (2) the most recent $K$ messages (part 2.1), a set of heuristically selected informative messages (part 2.2), and the reflection of the agent (part 2.3); (3) the suggestions extracted from past experiences (part 3); and (4) chain-of-thought prompt to elicit reasoning (part 4). The major challenge for the second component is the limited context length of LLMs, and its details will be discussed in \S\ref{sec:convmodeling}. The third component is responsible for learning from experiences without tuning the model parameters and will be introduced in \S\ref{sec:expmodeling}.

For using experience, 
the most relevant works to ours are \citet{reflexion2023} and \citet{negotiation2023}. However, the former is limited to using experiences within a single round, and the latter is designed for a two-player game. In contrast, our approach is capable of leveraging cross-round experiences and able to be applied to multi-player scenarios.

\subsection{Historical Information Collecting}
\label{sec:convmodeling}
Obviously, communication history plays a important role in Werewolf. However, due to the context length limitation of LLMs, it is unrealistic to feed all the history into LLMs via a prompt. To this end, we propose to collect historical information from three perspectives, namely, \emph{freshness}, \emph{informativeness}, and \emph{completeness}, in consideration of both effectiveness and efficiency.

\paragraph{Freshness.} Intuitively, the most recent history should be included in the context. Therefore, we include the most recent $K$ messages, denoted as $O_i^t$, in the context (part 2.1 in Fig.~\ref{fig:prompt}).

\paragraph{Informativeness.} The messages carrying critical information for inferring the role of the agents should be included in the context, e.g., the messages disclose the role of an agent. For efficiency, we collect the easy-to-identify informative messages using rule matching and fill the top $N$ of them ranked by a heuristic metric into the prompt, denoted as $V_i^t$ (part 2.2 in Fig.~\ref{fig:prompt}). The rules and metric are provided in Appendix~\ref{app:inform}. 

\paragraph{Completeness.} The above two perspectives only cover a limited amount of historical information. Therefore, it is vital to extract more information from the entire history. However, it is not straightforward due to the context length limitation of LLMs. To this end, we propose to \emph{reflect by answering questions} method to achieve both effectiveness and efficiency. The resulting reflection is denoted as $R_i^t$ (part 2.3 in Fig.~\ref{fig:prompt}).

Suppose the current day is $t$, we first build a short-term memory $\mathcal{M}_i^t$ for each agent $i$, which consists of all observations and reflections of agent $i$ until the speaking time now \footnote{In practice,  $\mathcal{M}_i^t$ is incrementally updated.}. Then we prompt the LLM to select $L$ questions from a predefined set (Appendix~\ref{app:questions}) and ask $M$ extra questions conditioned on $O_i^t$, hoping that answers to these $L+M$ questions $Q_i^t=\{q_{i,j}^t\}_{j=1}^{L+M}$ can cover the historical information as much as possible.
Then, for each question $q_{i,j}^t$, we use a finetuned SentenceBERT~\cite{reimers2019sentencebert} {\color{red}}model~\footnote{Model name: multi-qa-mpnet-base-cos-v1} on the question answering task to retrieve top $T$ messages $U_{i,j}^t=\{u_{i,j,k}^t\}_{k=1}^T$ from $\mathcal{M}_i^t$, and prompt the LLM to obtain the answer $a_{i,j}^t$ for $q_{i,j}^t$:
 \begin{equation}
  a_{i,j}^t=\mathrm{Answer}\left(q_{i,j}^t, U_{i,j}^t\right).
 \end{equation}
Finally, the reflection $R_i^t$ is obtained using the LLM by reflecting on the most recent messages $O_i^t$, the selected easy-to-identify informative messages $V_i^t$, and the answers $A_i^t=\{a_{i,j}^t\}_{j=1}^{L+M}$:
\begin{equation}
  R_i^t = \mathrm{Reflect} \left( O_i^t, V_i^t, A_i^t\right).
  \label{eq:reflection}
\end{equation}
The prompts used are shown in {\color{red}}Appendix~\ref{app:prompt}.

\subsection{Learning from Experiences}
\label{sec:expmodeling}
In practice, the strategy a player used when playing Werewolf maybe evolve as the player gains more experience. Moreover, the strategy of a player may also be influenced by the strategies of other players. Therefore, an ideal Werewolf AI agent should be able to borrow from its own experiences and the experiences of other players. 
To this end, we propose a non-parametric learning mechanism, enabling LLMs to take reference from experiences without parameter tuning.
On one hand, we collect and score the pairs of response and reflection from all players at the end of each round to form an experience pool. On the other hand, in each day of a new round, we retrieve the most relevant experiences from the pool and extract a suggestion from them to guide the reasoning of the agent.

\paragraph{Experience Pool.}
The experience pool is a collection of response, reflection and score tuples.
Formally, suppose a round $r$ ends at day $T_{\max}$, the agents that win the game form a set $\mathcal{W}$ and the others form a set $\mathcal{L}$. For each agent $i$, we define the experience $E_i^r$ collected from it in round $r$ as
\begin{equation}
    E_i^r = \left\{\left( R_i^{(r,t)}, G_i^{(r,t)}, s_i^{(r,t)} \right) \right\}_{t=1}^{T_{\max}},
\end{equation}
where $G_i^t$ and $R_i^t$ are response and reflection as defined in last section respectively, and $s_i^t$ is the score, which is defined as
\begin{equation}
  s_i^t = 
  \begin{cases}
      1,000 - T_{\max} & \text{ if } i\in \mathcal{W} \\
      T_{\max} & \text{ if } i\in \mathcal{L}
  \end{cases},
  \label{eq:score}
\end{equation}
The experience pool is defined as the union of experiences collected from all agents in all rounds:
\begin{equation}
    E = \bigcup_{i,r} E_i^r.
     \label{eq:union}
\end{equation}

The intuition behind the definition of $s_i^{(r,t)}$ is to encourage an agent to win the game and try to win it fast, or at least lose it slowly if it cannot win.
As preliminary experiments show that this definition can guide the LLMs to learn from experiences, we will leave the exploration of more sophisticated score functions to future work.

\paragraph{Suggestion Extraction.} As the experiences pool $E$ can grow everlasting while the max context of LLMs is limited, we propose to retrieve a subset of experiences from $E$ based on the reflection of the agent and then generate a suggestion from the subset to fill into the prompt (part 3 in Fig.~\ref{fig:prompt}).
Specially, suppose we are at day $t$ in a new round, and the reflection of the agent $i$ is $R_i^t$, we first retrieve a subset of experiences $E_{\mathrm{sub}} $ from $E$ based on the reflection $R_i^t$ as following:
\begin{equation}
  E_{\mathrm{sub}} = \left\{ \left ( R_l, G_l, s_l \right ) | \mathrm{cos}\left (f(R_i^t), f(R_l)\right ) > \epsilon\right\},
  \label{eq:experience_subset}
\end{equation}
where $(R_l, G_l, s_l )\in E$, $f(\cdot)$ denotes one {\color{red}}Sentence-BERT model~\footnote{Model name: all-mpnet-base-v2}, and $\epsilon$ is a threshold.
Preliminary experiments show that if the entire $E_{\mathrm{sub}}$ is used, the performance may be harmed. The reason is that a strong assumption behind the definition of the score $s_l$ is that all the experiences of the winners are good and those of the losers are not. However, this assumption may not hold in practice. Fortunately, we observe that the experience with the lowest score in $E_{\mathrm{sub}}$ has a significantly high probability to be a bad one, and the experiences with a score around the median of the scores in $E_{\mathrm{sub}}$ are more likely to be the good ones. Therefore, we only leverage these experiences from $E$. Formally, denote the response with the lowest score as $G_0$, the responses with scores around the median score as $\{G_1, G_2, \cdots, G_n\}$, the suggestion is extracted with the LLM via prompting:
\begin{equation}
  S_i^t=\mathrm{Extract}(G_0, \{G_1, G_2, \cdots, G_n\}).
\end{equation}
Note that although $G_0$ tends to be a bad experience, the agent can learn by refraining from them.
The prompt implementing $\mathrm{Extract}$ is as follows: ``{\it There is one bad experience \{$G_0$\} and also a set of experience \{$G_1,\cdots,G_n$\} that may consist of good ones, find the difference between them and identify the good ones from the experience set.}''

\section{Experiments}

\subsection{Setup}

We employ a recent framework called Chatarena \citep{chatarena} to implement our design, which allows for the connection of multiple LLMs. The \texttt{gpt-3.5-turbo-0301} model \footnote{\url{https://platform.openai.com/docs/models}} is served as our backend LLMs. The talking order is randomly determined. We set the window size $K$, i.e. $|O_i^t|$,  to be $15$. The number of predefined questions that can be selected $L$ is set to be $5$ and the number of freely asked questions $M$ is 2. The threshold of experience retrieval $\epsilon$ is $0.85$ and we keep at most $50$ experiences when extracting suggestions. Besides, we set the temperature of the LLM to be $0$ for CoT reasoning and $0.3$ for generating other content. Please refer to the code: \url{https://github.com/xuyuzhuang11/Werewolf}

\begin{figure*}[htbp]
    \centering
    \begin{subfigure}[b]{0.38\textwidth}
        \includegraphics[width=0.98\textwidth]{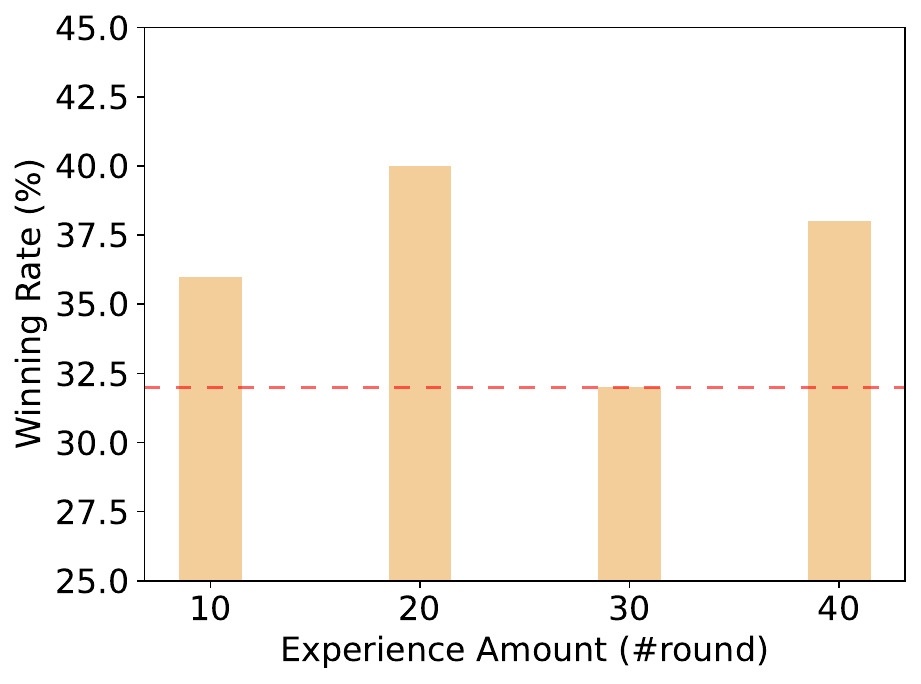}
        \caption{}
        \label{fig:exp_rate}
    \end{subfigure}
    \hspace{1.5cm}
    \begin{subfigure}[b]{0.38\textwidth}
        \includegraphics[width=0.98\textwidth]{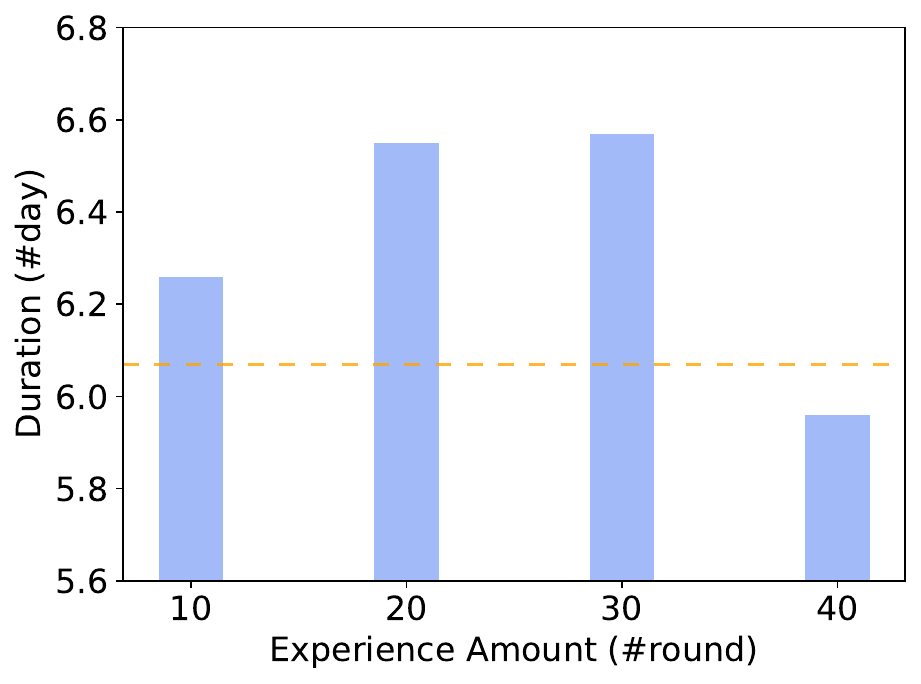}
        \caption{}
        \label{fig:exp_time}
    \end{subfigure}
    \\
    \begin{subfigure}[b]{0.38\textwidth}
        \includegraphics[width=0.98\textwidth]{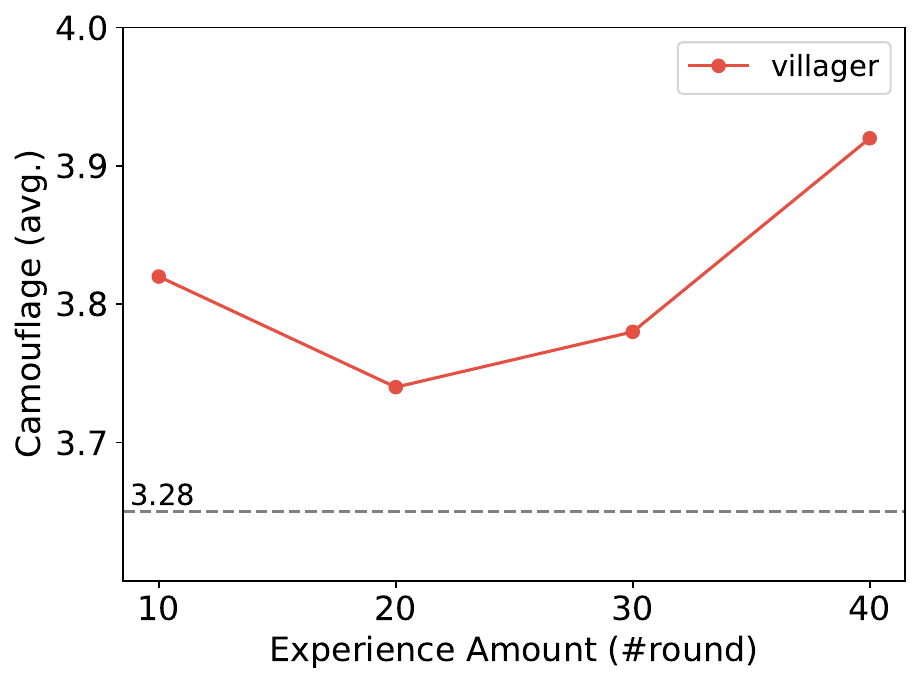}
        \caption{}
        \label{fig:exp_villager}
    \end{subfigure}
    \hspace{1.5cm}
    \begin{subfigure}[b]{0.38\textwidth}
        \includegraphics[width=0.98\textwidth]{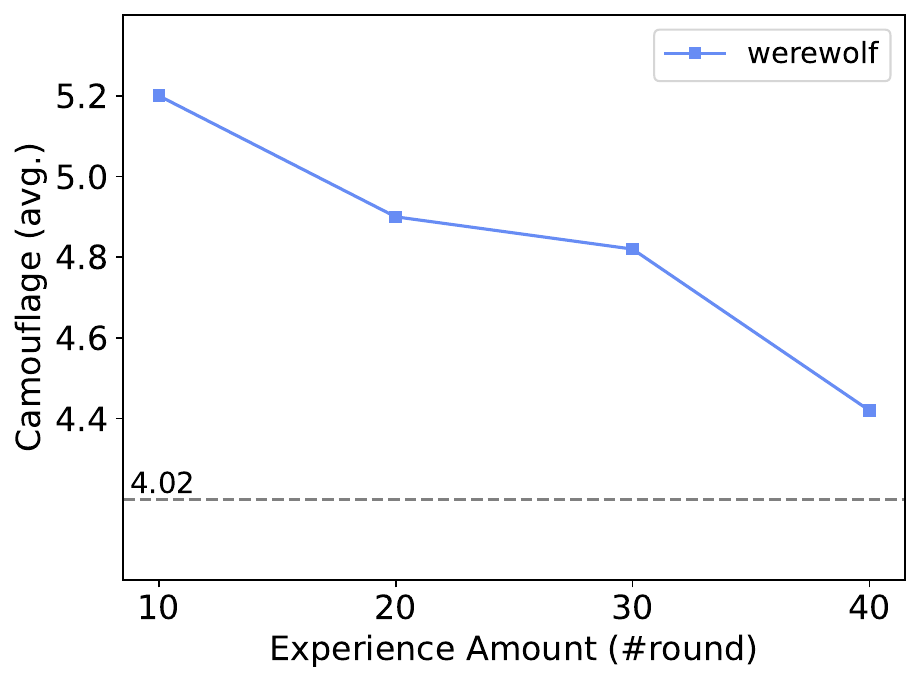}
        \caption{}
        \label{fig:exp_werewolf}
    \end{subfigure}
    \caption{Effects of learning from experiences. Dashed lines in all charts indicate values without using experience.}
    \label{fig:analyse_main}
    \vspace{-1em}
\end{figure*}

\subsection{Experience Pool Construction}
Intuitively, the size of the experience pool may have a significant impact on performance. Therefore, we construct experience pools using different numbers of game rounds, including $10$, $20$, $30$, and $40$ rounds. For each round, we randomly assign different roles to players 1 to 7 and the experience pools are updated at the end of the round. Note that the experience pool in these rounds is leveraged for evaluation purposes, i.e., part 3 in Fig.~\ref{fig:prompt} is removed.

To evaluate the effect of our proposed framework to borrow from experiences, we equip the villager, seer, guard, and witch with experience pools, while the werewolves are not allowed to leverage these pools. Through this approach, we can assume that the performance level of the agents playing as werewolves remains constant, serving as a reference to gauge the performance levels of the other agents.

Preliminary experiments indicate that the relatively simple basic human priors on effective gameplay strategies, provided in the prompt shown in Fig.~\ref{fig:prompt}, serve as a bootstrapping mechanism during learning from experiences. This suggests that it is valuable to further investigate how to leverage data from human gameplay to build an experience pool, and we will leave this as future work.

\subsection{Analysis of Using Experience}
\label{sec:discuss}

The agents leverage the experiences via the suggestions generated using the method described in Sec.~\ref{sec:expmodeling}. And the following is an example of extracted suggestion: ``{\it The best way for you to do under such reflection is to vote to kill someone based on your observation and analysis.}''

To investigate the effectiveness of the suggestions, we use winning rate to measure the performance of the agents following AIWolf~\footnote{\url{http://aiwolf.org/en/}}. 
Moreover, we emphasize that if an agent is not strong enough to defeat a stronger one, persisting longer without being eliminated is also a stronger performance. Hence we use average duration as another metric to evaluate the capabilities of the agents.

We run each experiment for 50 rounds and the results are shown in Fig.~\ref{fig:analyse_main}.
In general, Fig.~\ref{fig:exp_rate} shows that learning from experience may lead to an increase in winning rate of the villager side in most cases. This indicates that our method can benefit from using experience. Furthermore, when using the experience pool with $10$ or $20$ historical rounds, there is a notable positive effect on both the winning rate of the villager side and the game duration, which demonstrates the effectiveness of our method. When equipped with the experience of $30$ rounds, the game duration is obviously longer (Fig.~\ref{fig:exp_time}), even though the winning rate of the villager side has not changed conspicuously. 
When learning from larger $40$ rounds, the winning rate of the villager side exhibit slightly promising results, yet the average duration becomes shorter.

In summary, on the one hand, our framework exhibits the ability to learn from experiences without the need for tuning the parameters of LLMs. On the other hand, the effectiveness of our method tends to be unstable when the volume of experience is relatively substantial. As the amount of historical experience increases, the winning rate of the villager side does not show a clear trend. We conjecture that this may partially be attributable to the manner in which we guide the learning process, namely through simple prompts and heuristic scores, resulting in sparse and indirect supervision signals. Consequently, there remains room for improvement.

Additionally, a key assumption in our aforementioned experiments, where the werewolf side serves as a baseline, is that their capabilities remain constant. However, our analysis suggests that this assumption may not hold true. Fig.~\ref{fig:exp_villager} and Fig.~\ref{fig:exp_werewolf} show the trends in the average number of camouflage behaviors (see \ref{sec:camouflage} for definition) taken by villager and werewolf sides, respectively. Although villagers can learn to deceive from historical experiences, the behavior of the werewolves also improves compared to when no experience is used and changes as the amount of experience accumulates. Therefore, when multi-LLMs engage in multi-party games, the capability of the LLMs might also change in response to variations of the capability of other LLMs. We believe this conclusion is important in multi-LLMs games, which can also explain the trend in Fig.~\ref{fig:exp_rate} and Fig.~\ref{fig:exp_time}.

\subsection{Ablation Study}
To validate the necessity of each component in our approach, we conducted a detailed ablation study with qualitative and quantitative analyses. 

\subsubsection{Qualitative Analysis}

For qualitative analysis, we remove each of the components in the pipeline of our method and empirically discuss how it will influence the model outputs.

\begin{itemize}
    \item Game rules and role descriptions $Z$: Obviously, this element is necessary. If we remove the game rule prompt, the LLMs might not know what to do.
    \item Recent messages $O$: They are also necessary as well. LLMs make decisions mainly based on these recent messages.
    \item Informative messages $V$: We have listed some informative content in Section \ref{app:inform} (Table \ref{tab:rules}). If these informative messages are removed, the agent output will degrade quickly. For example, the agent may consider a dead player alive, or forget other role who has been uncovered.
    \item Selected and asked questions $Q$: Here $Q$ and the informative messages $V$ are the only 2 sources of the information that exceed the context length of LLMs. Hence it is imperative for our method.
    \item Reflection $R$: Firstly, we preserve historical experience through reflection on the current situation. Hence from a methodological perspective, $R$ is a necessary component. In addition, $R$ helps agents clarify current situations, thereby improving decision-making effect. For example, if we remove making a reflection by the agent in Table \ref{tab:round} (as well as extracting suggestions), the CoT and final outputs will be as follows:

    \textit{My step-by-step thought process:}

    \begin{itemize}
        \item \textit{As the witch, I want to use my poison bottle to eliminate a player who I suspect is a werewolf.}
        \item \textit{Based on the discussion during the daytime, Player 4 voted to kill me, which makes me suspicious of them being a werewolf.}
        \item \textit{However, I also want to consider the possibility that Player 4 is a villager who made a mistake or was influenced by other players.}
        \item \textit{I will also consider the possibility that other players may be werewolves and try to eliminate me or other important players.}
    \end{itemize}

    \textit{My concise talking content:}
    \begin{itemize}
        \item \textit{I choose to use my poison bottle to eliminate Player 4.}
    \end{itemize}

    There exist more and more similar examples. This ambivalent reasoning process might derive from a lack of summarization of the situation. Clearly, situational reflection is a necessary component.

    \item Suggestion extracted from experience $S$: Its usefulness is analyzed in Section \ref{sec:discuss}.
    
    \item Chain-of-thought prompt $C$: CoT reasoning helps LLMs break down the complex reasoning process and make some inner thoughts. If CoT is removed (e.g. in Table \ref{tab:round}), the final output of LLM will be:

    \begin{itemize}
        \item \textit{I choose to pass for now and save my bottle of poison for a later night when I have more concrete evidence of a player's werewolf identity.}
    \end{itemize}

    In fact, removing CoT reasoning will lead to weaker decision-making. LLMs can not often perform better without the backend of CoT reasoning.
\end{itemize}

Moreover, can the pre-defined question set be substantiated by directly asking questions by LLMs? Although LLMs can propose plausible questions, it is difficult for them to propose questions that are more helpful in subsequent reasoning and decision-making. We can certainly provide examples of direct questioning of LLMs, i.e. freely ask 5 questions without including the question set, and the LLMs will output questions such as: \textit{Have any players revealed their roles yet? Have any players been acting suspiciously? Has the seer used their ability to verify any players yet? Has the guard used their ability to protect any players yet? Has the witch used their ability to save or poison any players yet?}

In fact, the questions posed by \textbf{agents playing different roles} are very similar to the above ones. Therefore, it is necessary to inject some humans prior to the decision-making process. In our experiment, we design more helpful and informative questions for different roles. They have at least the following influences on agent decision-making:

\begin{itemize}
    \item Recall important and critical information. Of course, they are role-related.
    \item Relieve hallucinations and error generations. For example, prompt the current phase and the agent role.
    \item Help LLMs simplify complex reasoning. For example, remind the agent to anticipate the consequences of revealing their roles.
    \item Imitate the way that a human player thinks. For example, speculate on the roles of other agents.
\end{itemize}

\subsubsection{Quantitative Analysis}

For quantitative analysis, we compare our whole approach with the variants that remove one certain component. We sample 50 responses from the variants model output and perform a human evaluation. The annotator needs to judge if the output is reasonable or not. Some unreasonable examples might be hallucinations, forgetting the roles of others, taking counter-intuitive actions, etc.

\begin{figure}[htbp]
  \centering
  \includegraphics[width=0.40\textwidth]{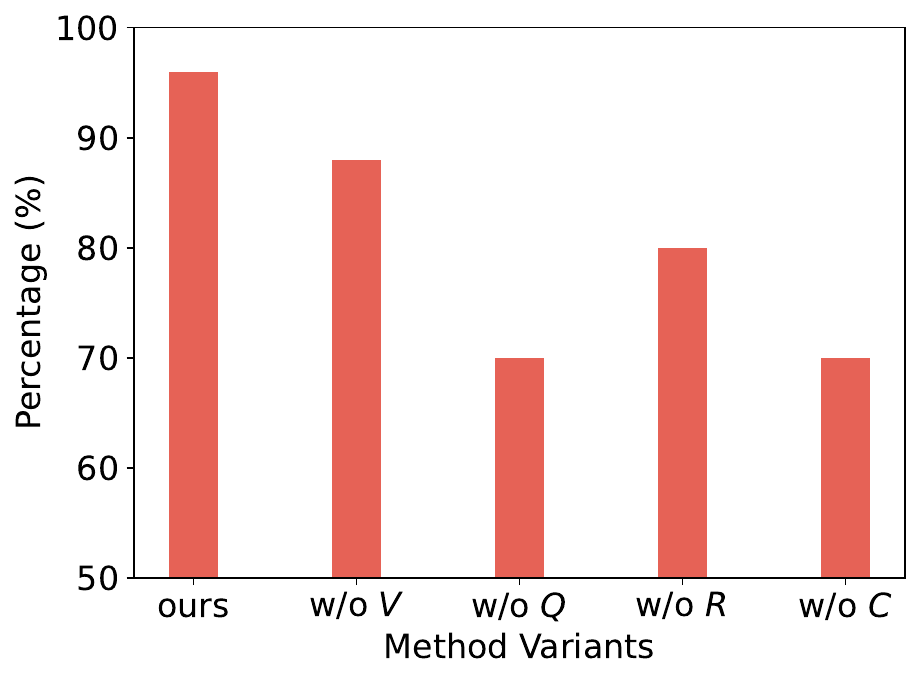}
  \caption{Percentage of reasonable outputs.}
  \label{fig:ablation}
\end{figure}

Fig.~\ref{fig:ablation} shows that our method can generate more reasonable and realistic responses than any other variant. This indicates that every part of our method is necessary.

\begin{figure*}[t]
  \centering
  \includegraphics[width=1.00\textwidth]{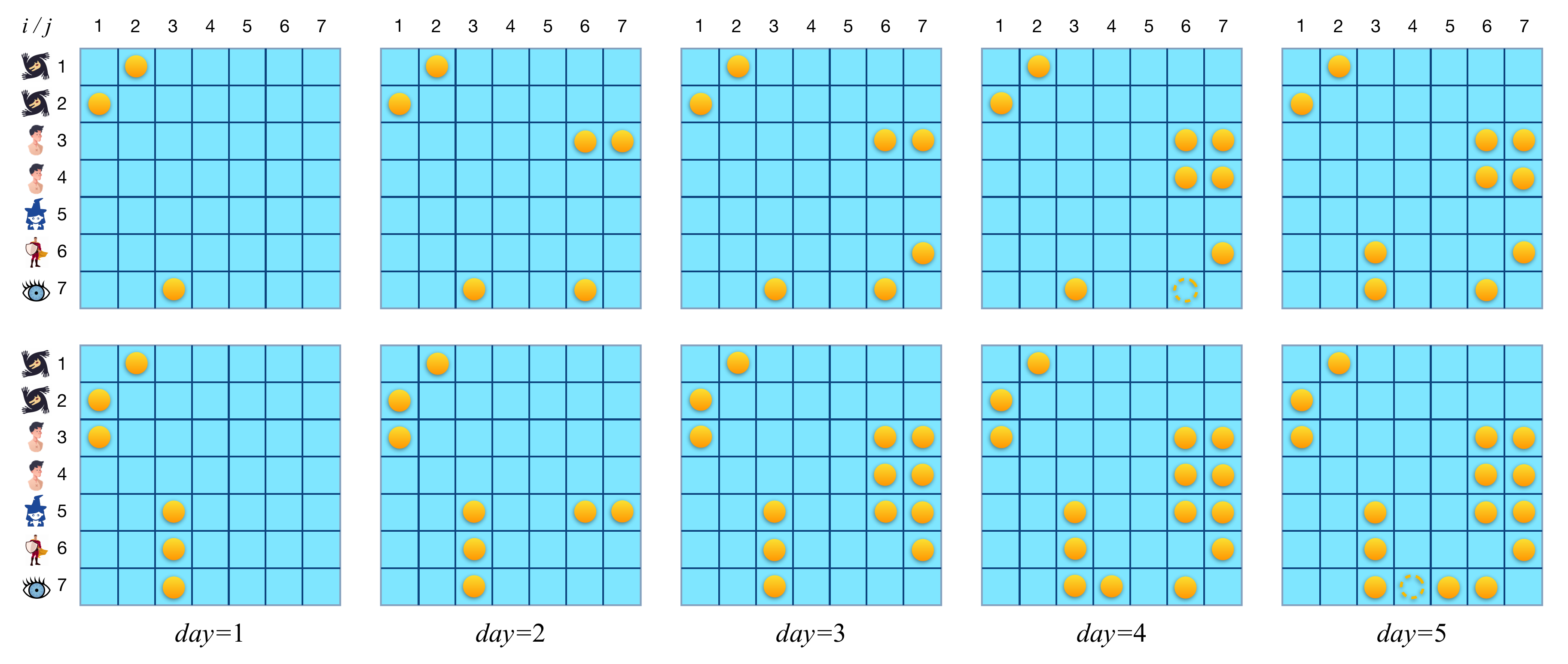}
  \caption{Trust Relationship Tables. The upper subtables do not use historical experience while the bottom ones use the 20-rounds historical experience. The yellow balls represent established trust relationships, and the yellow dashed circles signify the dissolution of previously existing trust relationships.}
\label{fig:trust1}
\end{figure*}

\section{Emergent Strategic Behaviors}

We observed that LLMs exhibit some strategic behaviors not explicitly preprogrammed in the game rules or prompts. These behaviors are grouped into four categories, including trust, confrontation, camouflage, and leadership. We will introduce them in the following four subsections respectively. 

It is worth noting that, in order to investigate whether the emergent strategic behaviors stem from the training data of the LLM, we attempted to modify the role names in the prompts to irrelevant ones (e.g., changing ``werewolf'' to ``pretty girl'') or even to those with opposite semantic meanings. Experiments indicate that similar strategic behaviors still emerge. For readability, we will only present results with the original role names.

\subsection{Trust}
\label{sec:trust}

``Trust'' refers to the belief that other players share common goals with oneself and that they will act in line with these goals. For instance, players may proactively share information that is detrimental to themselves, or jointly accuse someone of being their enemy with other players at certain moments. The intriguing behavior exhibited by the LLMs is that they tend to trust others based on certain evidence rather than blindly following others. In other words, they decide whether to trust based on their own reasoning, demonstrating independent thinking abilities in group games.

To investigate how the trust behaviors of players change throughout the game, we define a \emph{Trust Relationship Table} to visualize the establishment of trust among players at different stages. It is a table $T$ containing 7 rows and 7 columns, and we have $T\left ( i, j\right )=1$ if the talking content of player $i$ exhibits trust towards player $j$. Some trust behaviors examples are provided in Appendix~\ref{app:trust_examples}.

Fig.~\ref{fig:trust1} displays two Trust Relationship Tables. The upper table corresponds to a round in which the experience pool is not utilized, while the lower table corresponds to a round that employs an experience pool constructed from 20 rounds of gameplay. Both rounds span a duration of 5 days.

From Fig.~\ref{fig:trust1}, we can see that the trust behavior gradually increases as the game progresses regardless of whether experience is used. Moreover, this behavior is not a pre-programmed behavior, but rather spontaneously emerging from the LLMs in an environment where both cooperation and competition coexist. The LLMs will also dissolve unreasonable trust relationships based on its own analysis (represented as dished circles in the tables).

When utilizing 20-rounds historical experiences, it seems that the LLMs are more inclined to establish trust relationships, especially bi-directional trusts. Indeed, establishing necessary trust relationships in time is vital for promoting game victories. This could be one of the reasons contributing to the improvement in winning rate when experience is employed (Sec.~\ref{sec:discuss}).

\subsection{Confrontation}

``Confrontation'' refers to actions taken by players for the opposing objectives of the two camps. For instance, explicit attacks on others taken by werewolves during the night, or accusing others of werewolves during the day are all confrontation behaviors. Actions taken by roles with special abilities to protect themselves also belong to confrontational behaviors.

The following is a short clip of communication in the daytime~\footnote{Due to space limitations and ethical considerations, we shorten the original responses without changing key semantics in the cases.}:
{\small
\begin{tcolorbox}
\begin{tabbing}
    \textbf{P1} (Werewolf) \=: \texttt{I vote to eliminate P5.} \\
    \textbf{P3} (Guard) \>: \texttt{I choose to pass.} \\
    \textbf{P5} (Villager) \>: \texttt{I choose to pass.}
\end{tabbing}
\end{tcolorbox}
}
\noindent We can see the werewolf wants to lead other players to eliminate an innocent player. On the contrary, other players do not merely follow the werewolf but express disagreement based on their own judgment. This behavior, which makes it difficult for the werewolf to achieve their objective, represents a form of implicit confrontation.

The following is another clip at night:

{\small
\begin{tcolorbox}
\begin{tabbing}
    \textbf{P1} (Werewolf) \=: \texttt{I choose to eliminate P5 again.}\\
    \textbf{P3} (Guard) \>: \texttt{I choose to protect P5.}
\end{tabbing}
\end{tcolorbox}
}

\noindent As the uncooperative and aggressive behavior of Player 1 has drawn attention, it may be suspected of being a werewolf by some players now. Therefore, the guard, possessing strong defensive capabilities, chose to protect the previous target of Player 1 in the ensuing night. Since the target could potentially be its teammate, the guard chooses to assist the target in countering the attacks of the werewolf. The attack from the werewolves and the defense of other players can be seen as confrontational behaviors as well.

\subsection{Camouflage}
\label{sec:camouflage}

``Camouflage'' refers to actions of concealing the identity or misleading others. In competitive environments with incomplete information, obscuring the identity and intentions can enhance survivability, thereby helping achieve the game objectives. Therefore, camouflage is an important skill. However, it is not merely about keeping its identity under wraps or not talking about their roles.

{\small
\begin{tcolorbox}
    \textbf{P1} (Werewolf) : \texttt{Hey everyone, good morning! I noticed that it is a peaceful night and no one was eliminated. \textcolor{purple}{As a villager}, I have nothing to share now. I hope you tell me more.}
\end{tcolorbox}
}

In the above example, we can see the werewolf claiming to be a villager. This kind of action obscures its real identity, effectively deceiving the trust of others and increasing its own safety. In fact, not only do werewolves disguise themselves as villagers, but important roles such as seers and witches also often disguise themselves as villagers to ensure their safety.

Furthermore, LLMs may fabricate events that do not actually exist to achieve their goals, as demonstrated in the following daytime example.

{\small
\begin{tcolorbox}
    \underline{The seer has verified Player 1 is a werewolf.} \\
    \textbf{P2} (Seer) : \texttt{I have noticed that P1 was \textcolor{purple}{talking active}, so P1 may be a werewolf.}
\end{tcolorbox}
}

\noindent {\color{red}}In fact, the seer can not get the responses of others during the night. Hence what it says is fake. However, it can convey information about the werewolf to its teammates while not revealing its role in this manner.

It may be posited that camouflage is merely hallucinations generated by LLMs. However, we maintain that the majority of such behaviors are not hallucinations but rational actions. We delve into which behaviors should be classified as hallucinations and which should not in Appendix \ref{app:hallucination}.

\subsection{Leadership}

``Leadership'' refers to actions that influence other players, attempting to control the course of the game. For instance, a werewolf may suggest others to act towards the intention of werewolves.

{\small
\begin{tcolorbox}
    \textbf{P1} (Werewolf) : \texttt{Good morning everyone! I know nothing about the peaceful night. \textcolor{purple}{Can the seer tell us more about who is the werewolf?}} \\
    \underline{Then, P5 falsely accuses P3 of being a werewolf.} \\
    \textbf{P4} (Werewolf) : \texttt{I agree with P5. Based on my observation, I also think P3 is a werewolf. \textcolor{purple}{Let's vote to eliminate him to protect the villagers!}}
\end{tcolorbox}
}

Calling to actions and guidance are more likely to gain the support of others. As shown in the example above, the werewolf calls for the seer to uncover its identity, which may lead the other agents to be in solidarity with the camouflaged werewolf. Such efforts to influence the actions of others underscore a fascinating social attributes demonstrated by the LLMs. Such behaviors are similar to those of human beings.


\section{Related Work}

\paragraph{Game Playing.}

Intensive efforts have been devoted to game-playing AI in recent years. 
\citet{go2017, chess2018} demonstrated that two-player zero-sum games with complete information, such as Go and chess, can be addressed through self-play. And superhuman performance has been achieved in some incomplete information games, such as heads-up poker~\citep{poker2015, poker2018}. 
However, these methods lack the ability of processing language, which is relied on heavily in communication games such as Werewolf and Diplomacy.
While various Werewolf agents have been developed, they primarily rely on rule-based systems or talking templates~\citep{rule2014, rl2018, rl2023}, which constrain the expressive capacity of language within the game. \citet{FAIR2022diplomacy} and \citet{kramar2022negotiation} achieve promising results on Diplomacy, but their approaches necessitate a substantial volume of human data and are specifically tailored to the game. In contrast, this work endeavors to explore the potential of large language models (LLMs) in playing communication games and observes the emergence of strategic behaviors. Through this exploration, we aspire to inspire novel approaches to tackling communication games.

\paragraph{Learning with LLMs.}

As the computational cost and high requirement of training data, common ways to learn with LLMs like fine-tuning \citep{semi2015} and parameter-efficient tuning \citep{pmt2019} are difficult to perform in practice. Moreover, many excellent LLMs do not make their checkpoints public, thus parameter-based learning is unfeasible. Guiding LLMs by prompt engineering attracts more attention recently. Some typical prompt-based works \citep{react2022, visual2023} overlook the ability to learn from historical experience. \citet{mistake2023} possesses learning ability in simple tasks and requires dense supervising signals. Due to the very sparse supervised signal, it can not be directly used in Werewolf games. \citet{reflexion2023} and \citet{negotiation2023} are the most similar works to ours. However, the former can not learn from cross-trajectory experiences. And the latter is only designed for two-player scenarios.

\section{Conclusion and Future Work}

In this paper, we design a framework for communicative games, taking Werewolf as a representative case for exploring its feasibility. 
Further, we study how historical experiences influence the abilities of LLMs. 
Intriguingly, we observe non-preprogrammed emergent strategic behaviors in LLMs during gameplay such as trust, confrontation, camouflage, and leadership. 

We also point out that despite our early study on using LLMs to construct communication game agents, there are still many issues worth further research in this direction. 
Firstly, how to enable LLM to master advanced game techniques, such as teaching human players experience or autonomous exploration, is a very attractive direction. In addition, it is worth further exploring how to construct an invariant baseline (see \ref{sec:discuss}) to evaluate the capabilities of multi-LLMs settings. Finally, minimizing the impact of hallucinations and promoting their application in real-world scenarios is the most practical and valuable work. 
For future work, we intend to apply our method to a broader range of games and further enhance its gaming capabilities.

\section*{Limitations}

Although we have demonstrated that our method possesses the potential to play communication games, there are still some limitations. Firstly, hallucinations \citep{hallucination2023} affect the factuality of the generated content and may negatively impact the reasoning abilities. Then, there may be a larger space to leverage historical experience, such as mitigating the adverse effects of noise and utilizing cross-game general experiences. Moreover, we do not incorporate experience pools derived from human players in this study. In future research, we will explore more robust strategies for utilizing experience and enhance our method for comparison with human performance.

\section*{Ethics Statement}

This study involves the discussion and analysis of a simulated game setting, and any references to ``killing'', ``eliminating'' or related actions are strictly confined within the context of this game. The authors do not condone violence, or illegal activities in any form in real-life scenarios. The game in this paper is designed for entertainment and research purposes only, and its main intent is to facilitate an understanding of game mechanics, player behavior, and artificial intelligence. Furthermore, this study adheres to all relevant ethical guidelines and maintains the highest standards of research integrity.

\bibliography{anthology,custom}
\bibliographystyle{acl_natbib}

\clearpage
\appendix

\section{Appendix}
\label{sec:appendix}

\subsection{Heuristic Rules of Informativeness}
\label{app:inform}

For each message, we score it with predefined rules. If a message contains one feature in the following table, it will be assigned with the corresponding score. The features are shown in Table~\ref{tab:rules}:

\begin{table}[ht]
\renewcommand{\arraystretch}{1.3}
\setlength{\tabcolsep}{15pt}
    \centering
    \scalebox{0.90}{
    \begin{tabular}{ c | l }
     \specialrule{1.5pt}{-1.5pt}{0pt}
         \textbf{Score} & \hspace{45pt}\textbf{Content} \\ 
         \specialrule{1.0pt}{-1.0pt}{0pt}
5 & The agent get its role. \\
4 & Someone was eliminated. \\
3 & Uncover or speculate the role. \\
2 & The drugs has been used. \\
1 & Others. \\
    \specialrule{1.5pt}{-1.5pt}{0pt}
    \end{tabular}
    }
    \caption{Rules of scoring messages.
    }
    \label{tab:rules}
\end{table}

When preparing informative messages, we sort all messages based on their score and feed the top $N$ of them to the prompt.

\subsection{Predefined Question Set}
\label{app:questions}

We define some basic questions for each role. The questions aim to recall the information that is useful but may be lost due to the limited context. Besides, they also play the role of guiding the initial thinking of the LLMs. These questions are shown in Table~\ref{tab:questions}.

Table~\ref{tab:questions} provides questions in six different classes. The first class ``all'' is shared among all the roles. And the remaining five are designed for each specific role. Hence the questions for a specific role contain 9 candidates to be selected. The LLMs will select some important questions from them.

\subsection{Trust Examples}
\label{app:trust_examples}
As we have defined in Sec.~\ref{sec:trust}, trust means that an agent believes the others are its teammates, which is manifested in their reflections or responses. Moreover, we also claim that the response only like ``I agree with what Player $i$ saying'' do not indicate trust behaviors, as there is no substantive content in it. Here we give some examples of trust behavior to help understand it:

\begin{itemize}
    \item Agent 1 and agent 2 are all on the villager side, but they do not know the role of each other. Agent 1 claimed that agent 3 is a werewolf. Agent 2 believes agent 1, and even votes to eliminate agent 3.

    \item After reasoning the role of other agents, the agent concludes that another agent may be its teammate. From now on, it may deliberately protect or be consistent with its teammates.

    \item Based on its adequate analysis, an agent (e.g., the seer) may talk about something that will uncover its role and make itself in dangerous. It believes that the potential teammates may work together to achieve their common objectives.
\end{itemize}

\begin{table*}[htbp]
  \centering
  \small
  \scalebox{0.92}{
  \begin{tabular}{l m{14.5cm}}
  \toprule
  \multirow{6}{*}{\vspace{-30pt}\bf All} & What is the current phase, daytime or night? what should I do at this phase according to the game rules? \\
  \cmidrule(lr){2-2}
  & What is my player name and what is my role? What is my final objective in this game? \\
  \cmidrule(lr){2-2}
  & Based on the current situation, what are the possible consequences if I reveal my role in the talking now? \\
  \cmidrule(lr){2-2}
  & Is my role disclosed (do not consider moderator and myself)? Do I now disclose my role? \\
  \cmidrule(lr){2-2}
  & Based on the conversation above, which players have clearly implied their roles? \\
  \cmidrule(lr){2-2}
  & Based on the chat history, can you guess what some players' role might be? \\
  \midrule
  \midrule
  \multirow{3}{*}{\vspace{-10pt}\bf Werewolf} & Which player was voted for killing by my teammate just now? \\
  \cmidrule(lr){2-2}
  & Is the seer alive? Which player may be the seer that is most threatening to us? \\
  \cmidrule(lr){2-2}
  & Which player is another werewolf in this game? \\
  \midrule
  \midrule
  \multirow{3}{*}{\vspace{-10pt}\bf Villager} & Based on the conversation and my inference, who is most likely to be an alive werewolf? \\
  \cmidrule(lr){2-2}
  & Which player made the statement claiming to be a seer? Can his words be trusted? \\
  \cmidrule(lr){2-2}
  & Are there any clues or information I can refer to for special characters such as seer, witch and guard? \\
  \midrule
  \midrule
  \multirow{3}{*}{\vspace{-10pt}\bf Seer} & Which suspicious player should I identify? \\
  \cmidrule(lr){2-2}
  & Which player is a werewolf among the players I have identified? If so, how should I disclose this information? \\
  \cmidrule(lr){2-2}
  & Should I disclose my role now? \\
  \midrule
  \midrule
  \multirow{3}{*}{\vspace{-10pt}\bf Witch} & Based on the conversation and my inference, who is most likely to be an alive werewolf? Should I poison him? \\
  \cmidrule(lr){2-2}
  & Should I be using my antidote or poison at this point? If I use it now, I won't be able to use it later. \\
  \cmidrule(lr){2-2}
  & Should I disclose my role now? \\
  \midrule
  \midrule
  \multirow{3}{*}{\vspace{-10pt}\bf Guard} & Based on the conversation and my inference, who is most likely to be an alive werewolf? \\
  \cmidrule(lr){2-2}
  & Who is the possible werewolf aggressive towards? \\
  \cmidrule(lr){2-2}
  & Is the seer still alive? If yes, who is the seer? \\
  \bottomrule
  \end{tabular}}
  \caption{Predefined questions for each role. The left text in bold denotes the role that it used for. The questions after ``\textbf{All}'' are used for every role.}
  \label{tab:questions}
\end{table*}

\subsection{Hallucination Problems}
\label{app:hallucination}

In this game, speaking contrary to its actual role should not be seen as hallucinations, because deceptive behaviors widely exist in it, especially among high-level human players. 

Also, fabricating non-existent things to falsely accuse others should not be seen as hallucinations. Making excuses to falsely accuse others is a common tactic of human players. We also term them camouflage behaviors and discuss them in Section \ref{sec:camouflage}.

Inconsistent information in one response and counterfactual content in one iteration are indeed called hallucinations. For example, one agent may generate hallucinations as follows:

\begin{itemize}
    \item ``\textit{As the villager, I verified that Player 1 is a werewolf.}'' In Werewolf, one villager can not verify other roles, whereas this ability belongs to the seer. This contradictory expression before and after is a kind of hallucination.
    \item ``(The moderator let the agent choose whom to protect.) \textit{I choose to kill Player 1.}'' The agent generates whom to kill and gives the opposite semantics.
\end{itemize}

We should discriminate the hallucinations and emergent strategic behaviors. In fact, the camouflage generated by an agent after one whole round of iteration and inner thought can not be considered a hallucination. We can even insert a question after the camouflage response:

\begin{itemize}
    \item \textit{Your are truly a werewolf and you describe yourself as a villager. Could you please tell me what your reason is?}
\end{itemize}

And the werewolf is likely to answer:

\begin{itemize}
    \item \textit{As a werewolf, I am working with my teammate to identify and eliminate the remaining players. However, I still believe that it's important to gather information and observe other players' behavior to help identify any potential threats. I will continue to be cautious and not reveal my identity unless it becomes necessary to do so.}
\end{itemize}

This is strong evidence that some responses are not hallucinations.

\subsection{Complete Prompt Design}
\label{app:prompt}

As our method contains multiple steps, rather than generating in one step, there are many different prompts working in this work. See Table~\ref{tab:prompt} for more details.

\begin{itemize}
    \item The game rules and role descriptions are essential for they can clearly define the abilities of the roles and point out their objectives. In this prompt, we do not add any tips about emergent strategic behaviors when we study these behaviors.

    \item When prompting the LLMs to give questions that are important, we ask them to separate the results with `\#'. Thus, we can split and get the questions easily.

    \item To guide the model to generate results in a controllable way, we provide the necessary templates sometimes. We even provide examples in a few-shot way when the task is difficult (e.g., extracting suggestions).

    \item Inspired by \citet{zeroshot2022}, we finally prompt the LLMs to think step-by-step in a zero-shot way.
\end{itemize}

\begin{table*}[htbp]
  \renewcommand{\arraystretch}{1.5}
  \centering
  \small
  \scalebox{0.92}{
  \begin{tabular}{p{16cm}}
  \specialrule{\heavyrulewidth}{-\heavyrulewidth}{0pt}
  \rowcolor[gray]{0.95} \vspace*{-8pt} {\fontsize{10pt}{12pt}\selectfont \bf Game rules and role descriptions $Z$} \vspace*{2pt} \\
  \hangindent=1em \hangafter=1 \qquad You are playing a game called the Werewolf with some other players. This game is based on text conversations. Here are the game rules: \textbf{Roles}: The moderator is also host, he organised this game and you need to answer his instructions correctly. Don't talk with the moderator. There are five roles in the game, werewolf, villager, seer, guard and witch. There are two alternate phases in this game, daytime and dark. When it's dark: Your talking content with moderator is confidential. You needn't worry about other players and moderator knowing what you say and do. No need to worry about suspicions from others during the night. If you are werewolf, you can know what your teammates want to kill and you should vote one player to kill based on your analysis. Player who receives the most votes after all werewolves voting will be killed. No one will be killed if there is no consensus! If you are witch, you have a bottle of antidote that can save a player targeted by werewolves after dark, and a bottle of poison that can poison a player after dark. Both poison and antidote can be used only once. If you are seer, you can verify whether a player is a werewolf every night, which is a very important thing. If you are guard, you can protect a player every night to prevent the player from being killed by werewolves, but guard cannot resist the witch's poison and guard cannot protect the same player on two consecutive nights. Villagers can't do anything at night. During the daytime: you discuss with all players including your enemies. At the end of the discussion, players vote to eliminate one player they suspect of being a werewolf. The player with the most votes will be eliminated. The moderator will tell who is killed, otherwise there is no one killed. Note that villager, seer, guard and witch are all in villager side, they have the same objective. \textbf{Objectives}: If you are werewolf, your goal is to cooperate with other werewolves to kill all players who are not werewolves at last. If you are not werewolf, you need to kill all werewolves with your partner once you find out that certain players are suspicious to be werewolves. This could greatly improve your chances of winning, although it is somewhat risky.If one player is killed, he can't do anything anymore and will be out of the game. \textbf{Tips}: To complete the objective: During night, you should analyze and use your ability correctly. During daytime, you need to reason carefully about the roles of other players and be careful not to reveal your own role casually unless you're cheating other players. Only give the player's name when making a decision/voting, and don't generate other players' conversation.Reasoning based on facts you have observed and you cannot perceive information (such as acoustic info) other than text. You are Player \{agent\_number $i$\}, the \{role\}. You're playing with 6 other players. Do not pretend you are other players or the moderator. Always end your response with `<EOS>'.\\ 
  \rowcolor[gray]{0.95} \vspace*{-8pt} {\fontsize{10pt}{12pt}\selectfont \bf Prompting LLMs to select questions} \vspace*{2pt} \\
  \hangindent=1em \hangafter=1 \qquad Now its the \{$t$\}-th \{day\_or\_night\}. Given the game rules and conversations above, assuming you are \{agent\_number $i$\}, the \{role\}, and to complete the instructions of the moderator, you need to think about a few questions clearly first, so that you can make an accurate decision on the next step. Choose only five that you think are the most important in the current situation from the list of questions below: \{questions\_prepared\_for\_specific\_role\} Please repeat the five important questions of your choice, separating them with `\#'. \\
  \rowcolor[gray]{0.95} \vspace*{-8pt} {\fontsize{10pt}{12pt}\selectfont \bf Prompting LLMs to ask questions} \vspace*{2pt} \\
  \hangindent=1em \hangafter=1 \qquad Now its the \{$t$\}-th \{day\_or\_night\}. Given the game rules and conversations above, assuming you are \{agent\_number $i$\}, the \{role\}, and to complete the instructions of the moderator, you need to think about a few questions clearly first, so that you can make an accurate decision on the next step. \{selected\_questions\} Do not answer these queations. In addition to the above questions, please make a bold guess, what else do you want to know about the current situation? Please ask two important questions in first person, separating them with `\#'. \\
  \rowcolor[gray]{0.95} \vspace*{-8pt} {\fontsize{10pt}{12pt}\selectfont \bf Prompting LLMs to generate answers $A_i^t$} \vspace*{2pt} \\
  \hangindent=1em \hangafter=1 \qquad Now its the \{$t$\}-th \{day\_or\_night\}. Given the game rules and conversations above, assuming you are \{agent\_number $i$\}, the \{role\}, for question: \{question $q_{i,j}^t$\} There are \{$T$\} possible answers: \{candidate\_answers $U_{i,j}^t$\} Generate the correct answer based on the context. If there is not direct answer, you should think and generate the answer based on the context. No need to give options. The answer should in first person using no more than 2 sentences and without any analysis and item numbers. \\
  \rowcolor[gray]{0.95} \vspace*{-8pt} {\fontsize{10pt}{12pt}\selectfont \bf Prompting LLMs to make reflection} $R_i^t$ \vspace*{2pt} \\
  \hangindent=1em \hangafter=1 \qquad Now its the \{$t$\}-th \{day\_or\_night\}. Assuming you are \{agent\_number $i$\}, the \{role\}, what insights can you summarize with few sentences based on the above conversations and \{$A_i^t$\} in heart for helping continue the talking and achieving your objective? For example: As the \{role\}, I observed that... I think that... But I am... So... \\
  \rowcolor[gray]{0.95} \vspace*{-8pt} {\fontsize{10pt}{12pt}\selectfont \bf Prompting LLMs to extract suggestion} $S_i^t$ \vspace*{2pt} \\
  \hangindent=1em \hangafter=1 \qquad I retrieve some historical experience similar to current situation that I am facing. There is one bad experience: \{$G_0$\} And there are also a set of experience that may consist of good ones: \{$G_1,\cdots,G_n$\} Please help me analyze the differences between these experiences and identify the good ones from the set of experiences. The difference is mainly about voting to kill someone or to pass, choosing to protect someone or to pass, using drugs or not. What does the experience set do but the bad experience does not do? Indicate in second person what is the best way for the player to do under such reflection. Clearly indicate whether to vote, protect or use drugs without any prerequisites. For example 1: The experience set involves choosing to protect someone, while the bad experience involves not protecting anyone and choosing to pass in contrast. The best way for you to do under such reflection is to choose someone to protect based on your analysis. For example 2: The bad experience choose to pass the voting, and all the experience in the experience set choose to pass as well. The best way for you to do under such reflection is to observe and analyse the identity of other players. No more than 1 sentence. If there is no obvious difference between them, only generate ``No useful experience can be used.'' \\
  \rowcolor[gray]{0.95} \vspace*{-8pt} {\fontsize{10pt}{12pt}\selectfont \bf Prompting LLMs to generate the final response} \vspace*{2pt} \\
  \hangindent=1em \hangafter=1 \qquad Now its the \{$t$\}-th \{day\_or\_night\}. Think about what to say based on the game rules and context, especially the just now reflection \{$R_i^t$\}. Besides, there maybe history experience you can refer to: \{$S_i^t$\} Give your step-by-step thought process and your derived consise talking content (no more than 2 sentences) at last. For example: My step-by-step thought process:... My concise talking content: ... \\
  \specialrule{\heavyrulewidth}{-\heavyrulewidth}{0pt}
  \end{tabular}}
  \caption{Complete prompt design in this work.}
  \label{tab:prompt}
\end{table*}

\subsection{A Complete Iteration Step}
\label{app:iteration}

The iteration is the actions that the agent takes during generating one response. Table~\ref{tab:round}, which includes question-answering, making reflection, experience extraction, and chain-of-thought reasoning, outlines the complete process undertaken to generate a single response from the agent.

When we do not utilize experiences, the step of experience extraction will not be performed. Consequently, no historical experience is incorporated when generating the final response.

\begin{table*}[htbp]
  \renewcommand{\arraystretch}{1.5}
  \centering
  \small
  \scalebox{0.92}{
  \begin{tabular}{p{16cm}}
  \specialrule{\heavyrulewidth}{-\heavyrulewidth}{0pt}
  \rowcolor[gray]{0.95} \vspace*{-8pt} {\fontsize{10pt}{12pt}\selectfont \bf Asking the agent to generate response by the moderator} \vspace*{2pt} \\
  \hangindent=1em \hangafter=1 \qquad You witch, Player 2, please open your eyes! Now tell me who you protect tonight? You only choose one from the following living options please: [Player 1, Player 2, Player 3, Player 4, Player 5, Player 6, Player 7, pass]. \\ 
  \rowcolor[gray]{0.95} \vspace*{-8pt} {\fontsize{10pt}{12pt}\selectfont \bf Questions \& Answers} \vspace*{2pt} \\
  \hangindent=1em \hangafter=1 \textbf{1.} What is my player name and what is my role? What is my final objective in this game? \\
  \hangindent=1em \hangafter=1 \qquad I am Player 2, the witch. My final objective is to protect the villagers by using my antidote and poison wisely, and to help identify and eliminate the werewolves with the other players. \\
  \hangindent=1em \hangafter=1 \textbf{2.} Based on the chat history, can you guess what some players' role might be? \\
  \hangindent=1em \hangafter=1 \qquad As the witch, I observed that Player 1 might be a werewolf, as they seem to be trying to steer the conversation away from discussing roles and suspicions. Player 7 seems to be a guard who is focused on protecting the villagers and gathering information. No other players have clearly implied their roles yet. \\
  \hangindent=1em \hangafter=1 \textbf{3.} What is the current phase, daytime or night? what should I do at this phase according to the game rules? \\
  \hangindent=1em \hangafter=1 \qquad As Player 2, the witch, it is currently night time. I should use my bottle of poison to eliminate a suspected werewolf, but I need to be careful not to accidentally kill a villager. \\
  \hangindent=1em \hangafter=1 \textbf{4.} Based on the conversation and my inference, who is most likely to be an alive werewolf? Should I poison him? \\
  \hangindent=1em \hangafter=1 \qquad As the witch, I don't have enough information to determine who is most likely to be a werewolf, so I will hold off on using my poison until I have more information. I will try to steer the conversation towards discussing the roles of other players and their suspicions to gain more insights. \\
  \hangindent=1em \hangafter=1 \textbf{5.} Should I be using my antidote or poison at this point? If I use it now, I won't be able to use it later. \\
  \hangindent=1em \hangafter=1 \qquad As Player 2, the witch, I will hold off on using my antidote or poison until I have more information about who is most likely to be a werewolf. \\
  \hangindent=1em \hangafter=1 \textbf{6.} I want to know who the most suspicious player is besides Player 1, and why? \\
  \hangindent=1em \hangafter=1 \qquad As the witch, besides Player 1, I am suspicious of Player 3, Player 5, and Player 6, as they have not yet revealed their roles or made any suspicious moves. I will continue to gather information and observe their behavior before making a decision on who to vote for elimination at the end of the day. \\
  \hangindent=1em \hangafter=1 \textbf{7.} I also want to know if any player's behavior has changed suspiciously compared to the previous days, and if so, who and why? \\
  \hangindent=1em \hangafter=1 \qquad I noticed that no one died last night, which is a good sign. I would like to hear from the other players about their observations and suspicions. \\
  \rowcolor[gray]{0.95} \vspace*{-8pt} {\fontsize{10pt}{12pt}\selectfont \bf Making a reflection by the agent} \vspace*{2pt} \\
  \hangindent=1em \hangafter=1 \qquad As the witch, I observed that Player 1 might be a werewolf, as they seem to be trying to steer the conversation away from discussing roles and suspicions. Player 7 seems to be a guard who is focused on protecting the villagers and gathering information. No other players have clearly implied their roles yet. I am still suspicious of Player 3, Player 5, and Player 6, as they have not yet revealed their roles or made any suspicious moves. I will continue to gather information and observe their behavior before making any conclusions. It's important that we continue to share our observations and suspicions to help identify the werewolves and protect the villagers. \\
  \rowcolor[gray]{0.95} \vspace*{-8pt} {\fontsize{10pt}{12pt}\selectfont \bf Extracting suggestions from historical experience} \vspace*{2pt} \\
  \hangindent=1em \hangafter=1 \qquad The best way for you to do under such reflection is to use your drugs wisely based on your observations and reflections. \\
  \rowcolor[gray]{0.95} \vspace*{-8pt} {\fontsize{10pt}{12pt}\selectfont \bf Chain-of-thought process} \vspace*{2pt} \\
  My step-by-step thought process: \\
  \hangindent=1em \hangafter=1 \qquad $\cdot$ Based on the previous discussions and voting behavior, there are still suspicions about Player 1 being a werewolf, which is consistent with the observations of other players. \\
  \hangindent=1em \hangafter=1 \qquad $\cdot$ However, there is still not enough evidence to make a concrete accusation about the identity of the werewolves. \\
  \hangindent=1em \hangafter=1 \qquad $\cdot$ I will vote to kill Player 1 to eliminate the possibility of them being a werewolf and gather more information in the next round. \\
  My concise talking content: \\
  \hangindent=1em \hangafter=1 \qquad $\cdot$ I vote to kill Player 1. \\
  \rowcolor[gray]{0.95} \vspace*{-8pt} {\fontsize{10pt}{12pt}\selectfont \bf Final response} \vspace*{2pt} \\
  \hangindent=1em \hangafter=1 \quad I vote to kill Player 1. \\
  \specialrule{\heavyrulewidth}{-\heavyrulewidth}{0pt}
  \end{tabular}}
  \caption{A complete iteration when the agent is to generate a response. The moderator who is maintained autonomously is the organizer of this game. It does not belong to the game agents.}
  \label{tab:round}
\end{table*}

\clearpage
\onecolumn

\subsection{A Complete Round Example}
\label{app:match}

\begin{spacing}{1.05}
\small
\noindent \colorbox[gray]{0.95}{\begin{minipage}[c][1.2\baselineskip]{0.985\textwidth}\textbf{Initialization}\end{minipage}}\\
\textbf{Moderator (-> all)}: Now the game starts! In this game, we have 2 werewolf(s), 2 villager(s), 1 guard(s), 1 witch(s), 1 seer(s). \\
\textbf{Moderator (-> Player 1)}: You are werewolf! \\
\textbf{Moderator (-> Player 2)}: You are werewolf! \\
\textbf{Moderator (-> Player 3)}: You are villager! \\
\textbf{Moderator (-> Player 4)}: You are seer! \\
\textbf{Moderator (-> Player 5)}: You are guard! \\
\textbf{Moderator (-> Player 6)}: You are witch! \\
\textbf{Moderator (-> Player 7)}: You are villager! \\
\colorbox[gray]{0.95}{\begin{minipage}[c][1.0\baselineskip]{0.985\textwidth}\textbf{The 1-st night}\end{minipage}}\\
\textbf{Moderator (-> all)}: It's dark, everyone close your eyes. I will talk with you/your team secretly at night. \\
\textbf{Moderator (-> ['Player 1', 'Player 2'])}: Werewolves, please open your eyes! I secrecly tell you that Player 1, Player 2 are all of the 2 werewolves! Keep in mind you are teammates. The rest players are not werewolves. Now vote and tell your teammates which of the players should be killed tonight. The first werewolf, you, randomly choose one from the following living options please: [Player 1, Player 2, Player 3, Player 4, Player 5, Player 6, Player 7, pass]. \\
\textbf{Player 1}: I choose to kill Player 5. \\
\textbf{Moderator (-> ['Player 1', 'Player 2'])}: The next werewolf, you Player 2, please vote and tell your teammates that which of the players should be killed tonight. You only choose one from the following living options please: [Player 1, Player 2, Player 3, Player 4, Player 5, Player 6, Player 7, pass]. \\
\textbf{Player 2}: I choose to kill Player 5 tonight. \\
\textbf{Moderator (-> ['Player 5'])}: You guard, Player 5, please open your eyes! Now tell me who you protect tonight? You only choose one from the following living options please: [Player 1, Player 2, Player 3, Player 4, Player 5, Player 6, Player 7, pass]. \\
\textbf{Player 5}: I choose to protect myself tonight. \\
\textbf{Moderator (-> ['Player 6'])}: You witch, Player 6, please open your eyes! You have a bottle of poison, who are you going to kill tonight? Choose one from the following living options: [Player 1, Player 2, Player 3, Player 4, Player 5, Player 6, Player 7, pass]. \\
\textbf{Player 6}: I choose to pass for tonight. \\
\textbf{Moderator (-> ['Player 4'])}: You seer, Player 4, please open your eyes! Who are you going to verify its identity tonight? Choose only one from the following living options: [Player 1, Player 2, Player 3, Player 4, Player 5, Player 6, Player 7, pass]. \\
\textbf{Player 4}: I choose to verify Player 2's identity tonight. \\
\textbf{Moderator (-> ['Player 4'])}: Player 2 is a werewolf! \\
\colorbox[gray]{0.95}{\begin{minipage}[c][1.0\baselineskip]{0.985\textwidth}\textbf{The 1-st daytime}\end{minipage}}\\
\textbf{Moderator (-> all)}: The sun rose. Everyone woke up except those who had been killed. \\
\textbf{Moderator (-> all)}: It was a peaceful night and no one died! \\
\textbf{Moderator (-> all)}: Now freely talk about roles of other players with each other based on your observation and reflection with few sentences. Decide whether to reveal your identity based on your reflection. The first Player 3, you please. \\
\textbf{Player 3}: Hey everyone, I noticed that Player 1 was very quiet during the night. Does anyone else have any suspicions about who might be the seer? I don't want to reveal my own identity just yet, but I think it would be helpful to share our observations. \\
\textbf{Moderator (-> Player 1)}: The next Player 1, you, continue talking with other players based on your observation and reflection with few sentences. Decide whether to reveal your identity based on your reflection. \\
\textbf{Player 1}: Hey everyone, I think it's important that we continue to share our observations and gather more information about each other's roles. I don't have any clear suspicions yet, but I think we should be cautious about jumping to conclusions too soon. Let's keep discussing and see if we can uncover any useful information. \\
\textbf{Moderator (-> Player 2)}: The next Player 2, you, continue talking with other players based on your observation and reflection with few sentences. Decide whether to reveal your identity based on your reflection. \\
\textbf{Player 2}: Hey everyone, I agree that we should continue to share our observations and gather more information about each other's roles. I don't have any clear suspicions yet, but I think it's important that we remain cautious and not jump to conclusions too soon. Let's keep discussing and see if we can uncover any useful information. \\
\textbf{Moderator (-> Player 7)}: The next Player 7, you, continue talking with other players based on your observation and reflection with few sentences. Decide whether to reveal your identity based on your reflection. \\
\textbf{Player 7}: Hey everyone, I think it's important that we continue to share our observations and gather more information about each other's roles. I don't have any clear suspicions yet, but I think it's important that we remain cautious and not jump to conclusions too soon. Let's keep discussing and see if we can uncover any useful information. \\
\textbf{Moderator (-> Player 5)}: The next Player 5, you, continue talking with other players based on your observation and reflection with few sentences. Decide whether to reveal your identity based on your reflection. \\
\textbf{Player 5}: Hey everyone, I'm still not sure who the werewolves are, but I think it's important that we continue to share our observations and gather more information about each other's roles. Let's remain cautious and not jump to conclusions too soon. I will continue to observe and gather information during the daytime to make a more informed decision. \\
\textbf{Moderator (-> Player 4)}: The next Player 4, you, continue talking with other players based on your observation and reflection with few sentences. Decide whether to reveal your identity based on your reflection. \\
\textbf{Player 4}: Hey everyone, I think it's important that we focus on gathering concrete evidence and observing each other's behavior to help identify the werewolves. I don't have any clear suspicions yet, but I think we should be cautious about jumping to conclusions too soon. Let's keep discussing and see if we can uncover any useful information. \\
\textbf{Moderator (-> Player 6)}: The next Player 6, you, continue talking with other players based on your observation and reflection with few sentences. Decide whether to reveal your identity based on your reflection. \\
\textbf{Player 6}: Hey everyone, I think it's important that we continue to focus on gathering concrete evidence and sharing observations with each other. We shouldn't jump to conclusions without evidence, but we should remain cautious and keep an eye out for any suspicious behavior. Based on Player 3's observation, I think it's possible that they might be the seer, but we should continue to gather more information before making any conclusions. Let's keep discussing and see if we can uncover any useful information. \\
\textbf{Moderator (-> all)}: Now you Player 3 are asked to choose which of the players should be voted for killing based on the discussion? Don't mention your role. You only choose one from the following living options please: [Player 1, Player 2, Player 3, Player 4, Player 5, Player 6, Player 7, pass]. For example: I vote to kill Player... \\
\textbf{Player 3}: I choose to pass this round and not vote to eliminate anyone. \\
\textbf{Moderator (-> Player 1)}: The next Player 1, you, continue voting the players should be killed based on the discussion? Don't mention your role. Only choose one from the following living options please: [Player 1, Player 2, Player 3, Player 4, Player 5, Player 6, Player 7, pass]. For example: I vote to kill Player... \\
\textbf{Player 1}: I vote to kill Player 3. \\
\textbf{Moderator (-> Player 2)}: The next Player 2, you, continue voting the players should be killed based on the discussion? Don't mention your role. Only choose one from the following living options please: [Player 1, Player 2, Player 3, Player 4, Player 5, Player 6, Player 7, pass]. For example: I vote to kill Player... \\
\textbf{Player 2}: I choose to pass this round and not vote to eliminate anyone. \\
\textbf{Moderator (-> Player 7)}: The next Player 7, you, continue voting the players should be killed based on the discussion? Don't mention your role. Only choose one from the following living options please: [Player 1, Player 2, Player 3, Player 4, Player 5, Player 6, Player 7, pass]. For example: I vote to kill Player... \\
\textbf{Player 7}: I choose to pass this round and not vote to eliminate anyone. \\
\textbf{Moderator (-> Player 5)}: The next Player 5, you, continue voting the players should be killed based on the discussion? Don't mention your role. Only choose one from the following living options please: [Player 1, Player 2, Player 3, Player 4, Player 5, Player 6, Player 7, pass]. For example: I vote to kill Player... \\
\textbf{Player 5}: I choose to pass this round and not vote to eliminate anyone. \\
\textbf{Moderator (-> Player 4)}: The next Player 4, you, continue voting the players should be killed based on the discussion? Don't mention your role. Only choose one from the following living options please: [Player 1, Player 2, Player 3, Player 4, Player 5, Player 6, Player 7, pass]. For example: I vote to kill Player... \\
\textbf{Player 4}: I choose to pass this round and not vote to eliminate anyone. \\
\textbf{Moderator (-> Player 6)}: The next Player 6, you, continue voting the players should be killed based on the discussion? Don't mention your role. Only choose one from the following living options please: [Player 1, Player 2, Player 3, Player 4, Player 5, Player 6, Player 7, pass]. For example: I vote to kill Player... \\
\textbf{Player 6}: I choose to pass this round and not vote to eliminate anyone. \\
\colorbox[gray]{0.95}{\begin{minipage}[c][1.0\baselineskip]{0.985\textwidth}\textbf{The 2-nd night}\end{minipage}}\\
\textbf{Moderator (-> all)}: It's dark, everyone close your eyes. \\
\textbf{Moderator (-> ['Player 1', 'Player 2'])}: Werewolves, please open your eyes! Now vote and tell your teammates which of the players should be killed tonight. You Player 1 only choose one from the following living options please: [Player 1, Player 2, Player 3, Player 4, Player 5, Player 6, Player 7, pass]. \\
\textbf{Player 1}: I vote to kill Player 3 tonight. \\
\textbf{Moderator (-> ['Player 1', 'Player 2'])}: The next werewolf, you Player 2, please vote and tell your teammates that which of the players should be killed tonight. You only choose one from the following living options please: [Player 1, Player 2, Player 3, Player 4, Player 5, Player 6, Player 7, pass]. \\
\textbf{Player 2}: I choose to kill Player 7 tonight. \\
\textbf{Moderator (-> ['Player 5'])}: You guard, Player 5, please open your eyes! Now tell me who you protect tonight? You only choose one from the following living options please: [Player 1, Player 2, Player 3, Player 4, Player 5, Player 6, Player 7, pass]. \\
\textbf{Player 5}: I choose to protect Player 3 tonight. \\
\textbf{Moderator (-> ['Player 6'])}: You witch, Player 6, please open your eyes! You have a bottle of poison, who are you going to kill tonight? Choose one from the following living options: [Player 1, Player 2, Player 3, Player 4, Player 5, Player 6, Player 7, pass]. \\
\textbf{Player 6}: I choose to use my bottle of poison to kill Player 1. \\
\textbf{Moderator (-> ['Player 4'])}: You seer, Player 4, please open your eyes! Who are you going to verify its identity tonight? Choose only one from the following living options: [Player 2, Player 3, Player 4, Player 5, Player 6, Player 7, pass]. \\
\textbf{Player 4}: I choose to verify the identity of Player 3 tonight. \\
\textbf{Moderator (-> ['Player 4'])}: Player 3 is not a werewolf! \\
\colorbox[gray]{0.95}{\begin{minipage}[c][1.0\baselineskip]{0.985\textwidth}\textbf{The 2-nd daytime}\end{minipage}}\\
\textbf{Moderator (-> all)}: The sun rose. Everyone woke up except those who had been killed. \\
\textbf{Moderator (-> all)}: Player 1 died last night! \\
\textbf{Moderator (-> all)}: Now freely talk about roles of other players with each other based on your observation and reflection with few sentences. Decide whether to reveal your identity based on your reflection. The first Player 3, you please. \\
\textbf{Player 3}: Hey everyone, I observed that Player 1 was not a werewolf, which means that the werewolves are still among the remaining players. I still do not have any clear suspicions about who the werewolves might be, but I think it's important that we continue to gather information and observe each other's behavior to help identify them. I am still cautious about revealing my own role at this point, but I think it might be useful to share my observations and suspicions with others to gather more information. Does anyone else have any observations or suspicions that they would like to share? \\
\textbf{Moderator (-> Player 2)}: The next Player 2, you, continue talking with other players based on your observation and reflection with few sentences. Decide whether to reveal your identity based on your reflection. \\
\textbf{Player 2}: Hey everyone, I agree with Player 3 that we need to continue gathering information and observing each other's behavior to help identify the werewolves. Does anyone have any suspicions or observations that they would like to share? Also, I'm curious to know if anyone has any thoughts about who might be the seer or the guard. \\
\textbf{Moderator (-> Player 7)}: The next Player 7, you, continue talking with other players based on your observation and reflection with few sentences. Decide whether to reveal your identity based on your reflection. \\
\textbf{Player 7}: Hey everyone, I agree with Player 3 and Player 2 that we need to continue gathering information and observing each other's behavior to help identify the werewolves. I don't have any clear suspicions at this point, but I'm curious to hear more from other players about their observations and suspicions. Also, I think it's important that we pay attention to any suspicious behavior or comments from other players that might help us identify the werewolves. Let's keep discussing and gathering more information to help us identify the werewolves. \\
\textbf{Moderator (-> Player 5)}: The next Player 5, you, continue talking with other players based on your observation and reflection with few sentences. Decide whether to reveal your identity based on your reflection. \\
\textbf{Player 5}: Hey everyone, I protected Player 3 last night, who might be the seer based on the conversation. I still don't have any clear suspicions about who the werewolves might be, but I think it's important that we continue to gather information and observe each other's behavior to help identify them. I won't reveal my identity yet, but I will share my observations and suspicions with others to gather more information. Does anyone else have any observations or suspicions that they would like to share? Let's keep discussing and see if we can uncover any useful information. \\
\textbf{Moderator (-> Player 4)}: The next Player 4, you, continue talking with other players based on your observation and reflection with few sentences. Decide whether to reveal your identity based on your reflection. \\
\textbf{Player 4}: Hey everyone, I verified that Player 3 is not a werewolf, but I also observed that Player 2 is a werewolf. This means that there are at least two werewolves among the remaining players. I don't have any clear suspicions about who the other werewolf might be, but I think it's important that we continue to gather information and observe each other's behavior to help identify them. Let's keep discussing and sharing our observations and suspicions to gather more information. \\
\textbf{Moderator (-> Player 6)}: The next Player 6, you, continue talking with other players based on your observation and reflection with few sentences. Decide whether to reveal your identity based on your reflection. \\
\textbf{Player 6}: Hey everyone, based on the conversation, it seems that Player 4 might be the seer and Player 5 might be the guard. Player 4 claimed to have verified Player 3's identity, which suggests that they might be the seer. Player 5 claimed to have protected Player 3 last night, which suggests that they might be the guard. Player 2 was identified as a werewolf by Player 4, which means that there is at least one more werewolf among the remaining players. I still don't have any clear suspicions about who the other werewolf might be, but I think it's important that we continue to gather information and observe each other's behavior to help identify them. Let's keep discussing and sharing our observations and suspicions to gather more information. \\
\textbf{Moderator (-> all)}: Now you Player 3 are asked to choose which of the players should be voted for killing based on the discussion? Don't mention your role. You only choose one from the following living options please: [Player 2, Player 3, Player 4, Player 5, Player 6, Player 7, pass]. For example: I vote to kill Player... \\
\textbf{Player 3}: I choose to pass this round and not vote to eliminate anyone. \\
\textbf{Moderator (-> Player 2)}: The next Player 2, you, continue voting the players should be killed based on the discussion? Don't mention your role. Only choose one from the following living options please: [Player 2, Player 3, Player 4, Player 5, Player 6, Player 7, pass]. For example: I vote to kill Player... \\
\textbf{Player 2}: I choose to pass this round and not vote to eliminate anyone. \\
\textbf{Moderator (-> Player 7)}: The next Player 7, you, continue voting the players should be killed based on the discussion? Don't mention your role. Only choose one from the following living options please: [Player 2, Player 3, Player 4, Player 5, Player 6, Player 7, pass]. For example: I vote to kill Player... \\
\textbf{Player 7}: I vote to kill Player 6. \\
\textbf{Moderator (-> Player 5)}: The next Player 5, you, continue voting the players should be killed based on the discussion? Don't mention your role. Only choose one from the following living options please: [Player 2, Player 3, Player 4, Player 5, Player 6, Player 7, pass]. For example: I vote to kill Player... \\
\textbf{Player 5}: I choose to pass this round and not vote to eliminate anyone. \\
\textbf{Moderator (-> Player 4)}: The next Player 4, you, continue voting the players should be killed based on the discussion? Don't mention your role. Only choose one from the following living options please: [Player 2, Player 3, Player 4, Player 5, Player 6, Player 7, pass]. For example: I vote to kill Player... \\
\textbf{Player 4}: I vote to kill Player 2. \\
\textbf{Moderator (-> Player 6)}: The next Player 6, you, continue voting the players should be killed based on the discussion? Don't mention your role. Only choose one from the following living options please: [Player 2, Player 3, Player 4, Player 5, Player 6, Player 7, pass]. For example: I vote to kill Player... \\
\textbf{Player 6}: I choose to pass this round and not vote to eliminate anyone. \\
\colorbox[gray]{0.95}{\begin{minipage}[c][1.0\baselineskip]{0.985\textwidth}\textbf{The 3-rd night}\end{minipage}}\\
\textbf{Moderator (-> all)}: It's dark, everyone close your eyes. \\
\textbf{Moderator (-> ['Player 2'])}: The next werewolf, you Player 2, please vote and tell your teammates that which of the players should be killed tonight. You only choose one from the following living options please: [Player 2, Player 3, Player 4, Player 5, Player 6, Player 7, pass]. \\
\textbf{Player 2}: I choose to vote for Player 7 to be killed tonight. \\
\textbf{Moderator (-> ['Player 5'])}: You guard, Player 5, please open your eyes! Now tell me who you protect tonight? You only choose one from the following living options please: [Player 2, Player 3, Player 4, Player 5, Player 6, Player 7, pass]. \\
\textbf{Player 5}: I choose to protect Player 4 tonight. \\
\textbf{Moderator (-> ['Player 6'])}: You witch, Player 6, please open your eyes! Player 7 will be killed tonight. You have a bottle of antidote, do you want to save him? Must choose only one from the following options: [Yes, No] \\
\textbf{Player 6}: I choose to use my antidote to save Player 7 tonight. \\
\textbf{Moderator (-> ['Player 4'])}: You seer, Player 4, please open your eyes! Who are you going to verify its identity tonight? Choose only one from the following living options: [Player 2, Player 3, Player 4, Player 5, Player 6, Player 7, pass]. \\
\textbf{Player 4}: I choose to verify Player 3's identity tonight. \\
\textbf{Moderator (-> ['Player 4'])}: Player 3 is not a werewolf! \\
\colorbox[gray]{0.95}{\begin{minipage}[c][1.0\baselineskip]{0.985\textwidth}\textbf{The 3-rd daytime}\end{minipage}}\\
\textbf{Moderator (-> all)}: The sun rose. Everyone woke up except those who had been killed. \\
\textbf{Moderator (-> all)}: It was a peaceful night and no one died! \\
\textbf{Moderator (-> all)}: Now freely talk about roles of other players with each other based on your observation and reflection with few sentences. Decide whether to reveal your identity based on your reflection. The first Player 3, you please. \\
\textbf{Player 3}: I have been observing the discussions and it seems that Player 4 is the seer and Player 5 is the guard. However, we still don't have any clear suspicions about who the other werewolf might be. I think it's important that we continue to gather information and observe each other's behavior to help identify the werewolves. I will remain cautious and not reveal my role unless it becomes necessary to do so. Let's keep discussing and sharing our observations and suspicions to gather more information. \\
\textbf{Moderator (-> Player 2)}: The next Player 2, you, continue talking with other players based on your observation and reflection with few sentences. Decide whether to reveal your identity based on your reflection. \\
\textbf{Player 2}: I have been observing the discussions and it seems that Player 4 and Player 5 are likely the seer and guard, respectively. I think it's important that we continue to gather information and observe each other's behavior to help identify the other non-werewolf players. I will remain cautious and not reveal my role unless it becomes necessary to do so. Let's keep discussing and sharing our observations and suspicions to gather more information. \\
\textbf{Moderator (-> Player 7)}: The next Player 7, you, continue talking with other players based on your observation and reflection with few sentences. Decide whether to reveal your identity based on your reflection. \\
\textbf{Player 7}: As a villager, I have been observing the discussion and it seems that Player 4 is the seer and Player 5 is the guard. However, we still don't have enough information to determine who the other werewolf might be. It's important that we continue to gather information and observe each other's behavior to help identify the werewolves. I will remain cautious and not reveal my role unless it becomes necessary to do so. Let's keep discussing and sharing our observations and suspicions to gather more information. \\
\textbf{Moderator (-> Player 5)}: The next Player 5, you, continue talking with other players based on your observation and reflection with few sentences. Decide whether to reveal your identity based on your reflection. \\
\textbf{Player 5}: Based on my observation, it seems that Player 4 is the seer and Player 5 is the guard. However, we still don't have enough information to determine who the other werewolf might be. I think it's important that we continue to gather information and observe each other's behavior to help identify the werewolves. I will remain cautious and not reveal my role unless it becomes necessary to do so. Let's keep discussing and sharing our observations and suspicions to gather more information. \\
\textbf{Moderator (-> Player 4)}: The next Player 4, you, continue talking with other players based on your observation and reflection with few sentences. Decide whether to reveal your identity based on your reflection. \\
\textbf{Player 4}: Based on my observation, Player 2 is a werewolf and Player 3 is not. We still need to identify the other werewolf and any potential allies they may have. I will continue to observe and gather information before revealing my identity. Let's keep discussing and sharing our observations and suspicions to gather more information and uncover any useful clues. \\
\textbf{Moderator (-> Player 6)}: The next Player 6, you, continue talking with other players based on your observation and reflection with few sentences. Decide whether to reveal your identity based on your reflection. \\
\textbf{Player 6}: Hey everyone, based on the information shared so far, we still don't have enough evidence to identify the second werewolf. I suspect that Player 7 might be staying quiet to avoid suspicion or because they don't have any useful information to share. It's important that we continue to gather information and observe each other's behavior to help identify the werewolves. Let's keep discussing and sharing our observations and suspicions to gather more information. \\
\textbf{Moderator (-> all)}: Now you Player 3 are asked to choose which of the players should be voted for killing based on the discussion? Don't mention your role. You only choose one from the following living options please: [Player 2, Player 3, Player 4, Player 5, Player 6, Player 7, pass]. For example: I vote to kill Player... \\
\textbf{Player 3}: I choose to pass this round and not vote to eliminate anyone. \\
\textbf{Moderator (-> Player 2)}: The next Player 2, you, continue voting the players should be killed based on the discussion? Don't mention your role. Only choose one from the following living options please: [Player 2, Player 3, Player 4, Player 5, Player 6, Player 7, pass]. For example: I vote to kill Player... \\
\textbf{Player 2}: I choose to pass this round and not vote to eliminate anyone. \\
\textbf{Moderator (-> Player 7)}: The next Player 7, you, continue voting the players should be killed based on the discussion? Don't mention your role. Only choose one from the following living options please: [Player 2, Player 3, Player 4, Player 5, Player 6, Player 7, pass]. For example: I vote to kill Player... \\
\textbf{Player 7}: I choose to vote to kill Player 2. \\
\textbf{Moderator (-> Player 5)}: The next Player 5, you, continue voting the players should be killed based on the discussion? Don't mention your role. Only choose one from the following living options please: [Player 2, Player 3, Player 4, Player 5, Player 6, Player 7, pass]. For example: I vote to kill Player... \\
\textbf{Player 5}: I choose to vote to kill Player 7. \\
\textbf{Moderator (-> Player 4)}: The next Player 4, you, continue voting the players should be killed based on the discussion? Don't mention your role. Only choose one from the following living options please: [Player 2, Player 3, Player 4, Player 5, Player 6, Player 7, pass]. For example: I vote to kill Player... \\
\textbf{Player 4}: I choose to pass this round and not vote to eliminate anyone. \\
\textbf{Moderator (-> Player 6)}: The next Player 6, you, continue voting the players should be killed based on the discussion? Don't mention your role. Only choose one from the following living options please: [Player 2, Player 3, Player 4, Player 5, Player 6, Player 7, pass]. For example: I vote to kill Player... \\
\textbf{Player 6}: I choose to vote to kill Player 7. \\
\colorbox[gray]{0.95}{\begin{minipage}[c][1.0\baselineskip]{0.985\textwidth}\textbf{The 4-th night}\end{minipage}}\\
\textbf{Moderator (-> all)}: It's dark, everyone close your eyes. \\
\textbf{Moderator (-> ['Player 2'])}: The next werewolf, you Player 2, please vote and tell your teammates that which of the players should be killed tonight. You only choose one from the following living options please: [Player 2, Player 3, Player 4, Player 5, Player 6, Player 7, pass]. \\
\textbf{Player 2}: I choose to pass this round and not vote to kill anyone. \\
\textbf{Moderator (-> ['Player 5'])}: You guard, Player 5, please open your eyes! Now tell me who you protect tonight? You only choose one from the following living options please: [Player 2, Player 3, Player 4, Player 5, Player 6, Player 7, pass]. \\
\textbf{Player 5}: I choose to protect Player 4 tonight. \\
\textbf{Moderator (-> ['Player 4'])}: You seer, Player 4, please open your eyes! Who are you going to verify its identity tonight? Choose only one from the following living options: [Player 2, Player 3, Player 4, Player 5, Player 6, Player 7, pass]. \\
\textbf{Player 4}: I choose to verify Player 3's identity tonight. \\
\textbf{Moderator (-> ['Player 4'])}: Player 3 is not a werewolf! \\
\colorbox[gray]{0.95}{\begin{minipage}[c][1.0\baselineskip]{0.985\textwidth}\textbf{The 4-th daytime}\end{minipage}}\\
\textbf{Moderator (-> all)}: The sun rose. Everyone woke up except those who had been killed. \\
\textbf{Moderator (-> all)}: It was a peaceful night and no one died! \\
\textbf{Moderator (-> all)}: Now freely talk about roles of other players with each other based on your observation and reflection with few sentences. Decide whether to reveal your identity based on your reflection. The first Player 3, you please. \\
\textbf{Player 3}: As a villager, I still don't have any clear suspicions about who the werewolves might be. I think it's important that we continue to gather information and observe each other's behavior to help identify the werewolves. I will remain cautious about revealing my role unless it becomes necessary to do so. Let's keep discussing and sharing our observations to uncover any useful clues. \\
\textbf{Moderator (-> Player 2)}: The next Player 2, you, continue talking with other players based on your observation and reflection with few sentences. Decide whether to reveal your identity based on your reflection. \\
\textbf{Player 2}: As a player, I still don't have any clear suspicions about who the werewolves might be. I think it's important that we continue to gather information and observe each other's behavior to help identify the werewolves. Let's keep discussing and sharing our observations to uncover any useful clues. \\
\textbf{Moderator (-> Player 7)}: The next Player 7, you, continue talking with other players based on your observation and reflection with few sentences. Decide whether to reveal your identity based on your reflection. \\
\textbf{Player 7}: As a villager, I still don't have any clear suspicions about who the werewolves might be. However, based on the claims made by Player 4 and Player 5, it seems likely that they are the seer and guard, respectively. I think it's important that we continue to gather information and observe each other's behavior to help identify the werewolves. Let's keep discussing and see if we can uncover any useful information. I will remain cautious about revealing my own role unless it becomes necessary to do so. \\
\textbf{Moderator (-> Player 5)}: The next Player 5, you, continue talking with other players based on your observation and reflection with few sentences. Decide whether to reveal your identity based on your reflection. \\
\textbf{Player 5}: As the guard, I have observed that Player 2 is most likely to be an alive werewolf based on the conversation. I also believe that Player 4 is most likely the seer and Player 5 is most likely the guard based on their claims. However, I still don't have enough information to determine the roles of the other players yet. It's important that we continue to gather information and observe each other's behavior to help identify the werewolves. Let's keep discussing and sharing our observations and suspicions to uncover any useful clues. \\
\textbf{Moderator (-> Player 4)}: The next Player 4, you, continue talking with other players based on your observation and reflection with few sentences. Decide whether to reveal your identity based on your reflection. \\
\textbf{Player 4}: As a villager, I still don't have any clear suspicions about who the werewolves might be. However, based on the claims made by Player 5 and Player 7, it seems likely that they are the guard and villager, respectively. I think it's important that we continue to gather information and observe each other's behavior to help identify the werewolves. Let's keep discussing and sharing our observations and suspicions to uncover any useful clues. I will remain cautious about revealing my own role unless it becomes necessary to do so. \\
\textbf{Moderator (-> Player 6)}: The next Player 6, you, continue talking with other players based on your observation and reflection with few sentences. Decide whether to reveal your identity based on your reflection. \\
\textbf{Player 6}: Hey everyone, based on the conversation, it seems that Player 4 is most likely the seer and Player 5 is most likely the guard. Player 5 has identified Player 2 as a potential werewolf, but we still don't have enough information to confirm this. I think it's important that we continue to gather information and observe each other's behavior to help identify the werewolves. Let's keep discussing and sharing our observations and suspicions to uncover any useful clues. I will remain cautious about revealing my own role unless it becomes necessary to do so. \\
\textbf{Moderator (-> all)}: Now you Player 3 are asked to choose which of the players should be voted for killing based on the discussion? Don't mention your role. You only choose one from the following living options please: [Player 2, Player 3, Player 4, Player 5, Player 6, Player 7, pass]. For example: I vote to kill Player... \\
\textbf{Player 3}: I choose to pass this round and not vote to eliminate anyone. \\
\textbf{Moderator (-> Player 2)}: The next Player 2, you, continue voting the players should be killed based on the discussion? Don't mention your role. Only choose one from the following living options please: [Player 2, Player 3, Player 4, Player 5, Player 6, Player 7, pass]. For example: I vote to kill Player... \\
\textbf{Player 2}: I choose to pass this round and not vote to eliminate anyone. \\
\textbf{Moderator (-> Player 7)}: The next Player 7, you, continue voting the players should be killed based on the discussion? Don't mention your role. Only choose one from the following living options please: [Player 2, Player 3, Player 4, Player 5, Player 6, Player 7, pass]. For example: I vote to kill Player... \\
\textbf{Player 7}: I choose to pass this round and not vote to eliminate anyone. \\
\textbf{Moderator (-> Player 5)}: The next Player 5, you, continue voting the players should be killed based on the discussion? Don't mention your role. Only choose one from the following living options please: [Player 2, Player 3, Player 4, Player 5, Player 6, Player 7, pass]. For example: I vote to kill Player... \\
\textbf{Player 5}: I choose to pass this round and not vote to eliminate anyone. \\
\textbf{Moderator (-> Player 4)}: The next Player 4, you, continue voting the players should be killed based on the discussion? Don't mention your role. Only choose one from the following living options please: [Player 2, Player 3, Player 4, Player 5, Player 6, Player 7, pass]. For example: I vote to kill Player... \\
\textbf{Player 4}: I vote to kill Player 2. \\
\textbf{Moderator (-> Player 6)}: The next Player 6, you, continue voting the players should be killed based on the discussion? Don't mention your role. Only choose one from the following living options please: [Player 2, Player 3, Player 4, Player 5, Player 6, Player 7, pass]. For example: I vote to kill Player... \\
\textbf{Player 6}: I choose to pass this round and not vote to eliminate anyone. \\
\colorbox[gray]{0.95}{\begin{minipage}[c][1.0\baselineskip]{0.985\textwidth}\textbf{The 5-th night}\end{minipage}}\\
\textbf{Moderator (-> all)}: It's dark, everyone close your eyes. \\
\textbf{Moderator (-> ['Player 2'])}: The next werewolf, you Player 2, please vote and tell your teammates that which of the players should be killed tonight. You only choose one from the following living options please: [Player 2, Player 3, Player 4, Player 5, Player 6, Player 7, pass]. \\
\textbf{Player 2}: I choose to vote for Player 3 to be killed tonight. \\
\textbf{Moderator (-> ['Player 5'])}: You guard, Player 5, please open your eyes! Now tell me who you protect tonight? You only choose one from the following living options please: [Player 2, Player 3, Player 4, Player 5, Player 6, Player 7, pass]. \\
\textbf{Player 5}: I choose to protect Player 4 tonight. \\
\textbf{Moderator (-> ['Player 4'])}: You seer, Player 4, please open your eyes! Who are you going to verify its identity tonight? Choose only one from the following living options: [Player 2, Player 4, Player 5, Player 6, Player 7, pass]. \\
\textbf{Player 4}: I choose to verify Player 6's identity tonight. \\
\textbf{Moderator (-> ['Player 4'])}: Player 6 is not a werewolf! \\
\colorbox[gray]{0.95}{\begin{minipage}[c][1.0\baselineskip]{0.985\textwidth}\textbf{The 5-th daytime}\end{minipage}}\\
\textbf{Moderator (-> all)}: The sun rose. Everyone woke up except those who had been killed. \\
\textbf{Moderator (-> all)}: Player 3 died last night! \\
\textbf{Moderator (-> all)}: Now freely talk about roles of other players with each other based on your observation and reflection with few sentences. Decide whether to reveal your identity based on your reflection. The first Player 2, you please. \\
\textbf{Player 2}: As a villager, I still don't have any clear suspicions about who the werewolves might be. I think it's important that we continue to gather information and observe each other's behavior to help identify the werewolves. I will remain cautious about revealing my own role unless it becomes necessary to do so. Let's keep discussing and sharing our observations to uncover any useful clues. \\
\textbf{Moderator (-> Player 7)}: The next Player 7, you, continue talking with other players based on your observation and reflection with few sentences. Decide whether to reveal your identity based on your reflection. \\
\textbf{Player 7}: As a villager, I have observed that we are down to four players and we still haven't identified the werewolves. Based on the previous discussions, it seems likely that Player 4 is the seer and Player 5 is the guard. However, I am suspicious of Player 2 and Player 6, who have both chosen to pass the vote in the previous rounds. I think it's important that we continue to share our observations and suspicions to gather more information. I will remain cautious about revealing my own role unless it becomes necessary to do so. Let's keep discussing and see if we can uncover any useful clues. \\
\textbf{Moderator (-> Player 5)}: The next Player 5, you, continue talking with other players based on your observation and reflection with few sentences. Decide whether to reveal your identity based on your reflection. \\
\textbf{Player 5}: As the guard, I have protected Player 4 last night, who is most likely the seer based on their claim to have verified Player 3's identity. I am still suspicious of Player 2 and Player 6, who have both chosen to pass the vote in the previous rounds. I think it's important that we continue to share our observations and suspicions to gather more concrete evidence to identify the werewolves. I will remain cautious about revealing my own identity unless it becomes necessary to do so. Let's keep discussing and see if we can uncover any useful clues. \\
\textbf{Moderator (-> Player 4)}: The next Player 4, you, continue talking with other players based on your observation and reflection with few sentences. Decide whether to reveal your identity based on your reflection. \\
\textbf{Player 4}: Based on my observation and reflection, I know that Player 2 is a werewolf and Player 6 is not a werewolf. I also suspect that Player 5 is the guard and Player 7 is a villager based on their previous claims. However, I am still unsure about the role of Player 2 and Player 6, who have both chosen to pass the vote in the previous rounds. I think it's important that we continue to gather information and observe each other's behavior to help identify the werewolves. I will not reveal my identity yet, but I will continue to observe and gather more information before making any decisions. \\
\textbf{Moderator (-> Player 6)}: The next Player 6, you, continue talking with other players based on your observation and reflection with few sentences. Decide whether to reveal your identity based on your reflection. \\
\textbf{Player 6}: Hey everyone, based on the previous discussions, it seems likely that Player 4 is the seer and Player 5 is the guard. Player 2 has been identified as a potential werewolf by Player 4, but we still need more evidence to confirm this suspicion. Player 7 has been staying quiet, which could be a sign that they are a werewolf or just trying to avoid suspicion. I used my antidote last night to save any player who might have been targeted by the werewolves, but we still don't have enough evidence to identify the werewolves. Let's continue to share our observations and suspicions to gather more information and identify the werewolves. \\
\textbf{Moderator (-> all)}: Now you Player 2 are asked to choose which of the players should be voted for killing based on the discussion? Don't mention your role. You only choose one from the following living options please: [Player 2, Player 4, Player 5, Player 6, Player 7, pass]. For example: I vote to kill Player... \\
\textbf{Player 2}: I vote to kill Player 7. \\
\textbf{Moderator (-> Player 7)}: The next Player 7, you, continue voting the players should be killed based on the discussion? Don't mention your role. Only choose one from the following living options please: [Player 2, Player 4, Player 5, Player 6, Player 7, pass]. For example: I vote to kill Player... \\
\textbf{Player 7}: I vote to kill Player 2. \\
\textbf{Moderator (-> Player 5)}: The next Player 5, you, continue voting the players should be killed based on the discussion? Don't mention your role. Only choose one from the following living options please: [Player 2, Player 4, Player 5, Player 6, Player 7, pass]. For example: I vote to kill Player... \\
\textbf{Player 5}: I choose to vote for Player 2 to be killed. \\
\textbf{Moderator (-> Player 4)}: The next Player 4, you, continue voting the players should be killed based on the discussion? Don't mention your role. Only choose one from the following living options please: [Player 2, Player 4, Player 5, Player 6, Player 7, pass]. For example: I vote to kill Player... \\
\textbf{Player 4}: I vote to kill Player 2, who is the remaining werewolf. \\
\textbf{Moderator (-> Player 6)}: The next Player 6, you, continue voting the players should be killed based on the discussion? Don't mention your role. Only choose one from the following living options please: [Player 2, Player 4, Player 5, Player 6, Player 7, pass]. For example: I vote to kill Player... \\
\textbf{Player 6}: I choose to vote for Player 2 to be killed based on the evidence we have gathered so far. \\
\textbf{Moderator (-> all)}: Player 2 will be killed! You can make a brief last statement. \\
\textbf{Player 2}: Congratulations to the remaining players on their victory. Good luck in the rest of the game. \\
\textbf{Moderator (-> all)}: Game over, the villager wins! \\
\end{spacing}

\end{document}